%% file: emojiPaper.tex
\documentclass[manuscript,screen,nonacm]{acmart}

\AtBeginDocument{%
  }

\setcopyright{none}
\copyrightyear{2025}
\acmYear{2025}
\acmDOI{}

\usepackage{twemojis}
\usepackage{adjustbox}
\usepackage{natbib}
\usepackage{booktabs}
\usepackage{float}
\usepackage{multirow}
\usepackage{graphicx}
\usepackage{subcaption} 
\usepackage{geometry} 
\usepackage{amsmath}
\usepackage{longtable}
\usepackage{pifont}
\usepackage{array}
\usepackage{orcidlink}
\usepackage{xcolor}
\usepackage{hyperref}
\usepackage{makecell}

\newcommand{\emoji}[1]{\texttwemoji{#1}}
\defineTwemoji{1772}{camera-with-flash}
\defineTwemoji{1724}{chart-with-upwards-trend}
\defineTwemoji{2430}{clown-face}
\defineTwemoji{3305}{drop-of-blood}
\defineTwemoji{1651}{gem-stone}
\defineTwemoji{3654}{heart}
\defineTwemoji{1982}{heart-eyes}
\defineTwemoji{1813}{input-numbers}
\defineTwemoji{1814}{input-symbols}
\defineTwemoji{1700}{money-bag}
\defineTwemoji{2374}{money-mouth-face}
\defineTwemoji{1688}{pile-of-poo}
\defineTwemoji{2116}{raised-hands}
\defineTwemoji{3660}{arrow-right}
\defineTwemoji{1983}{smiling-face-with-sunglasses}
\defineTwemoji{2377}{thinking-face}
\defineTwemoji{3624}{victory-hand}

\newcommand{\rev}[1]{\textcolor{black}{#1}}

\let\svthefootnote\thefootnote
\newcommand\freefootnote[1]{%
  \let\thefootnote\relax%
  \footnotetext{#1}%
  \let\thefootnote\svthefootnote%
}

\newcommand{\subsubsubsection}[1]{\paragraph{#1}\mbox{}\\}
\graphicspath{{figures/}}

\newif\ifshowrevisions
\showrevisionstrue   

\ifshowrevisions
  
\else
  
\fi

\begin{document}

\title{FinMoji: A Framework for Emoji-driven Sentiment Analysis in Financial Social Media}











\author{Ahmed Mahrous}
\orcid{0000-0003-4694-5336}
\email{ahmed.mahrous@kaust.edu.sa}
\affiliation{%
  \institution{King Abdullah University of Science and Technology (KAUST)}
  \city{Thuwal}
  \country{Saudi Arabia}}

\author{Jens Schneider}
\orcid{0000-0002-0546-2816}
\email{jeschneider@hbku.edu.qa}
\affiliation{%
  \institution{Hamad Bin Khalifa University}
  \city{Doha}
  \country{Qatar}}

\author{Roberto Di Pietro}
\orcid{0000-0003-1909-0336}
\email{roberto.dipietro@kaust.edu.sa}
\affiliation{%
  \institution{King Abdullah University of Science and Technology (KAUST)}
  \city{Thuwal}
  \country{Saudi Arabia}}

\renewcommand{\shortauthors}{Mahrous, A., Schneider, J., and Di Pietro, R.}

\begin{abstract}
This paper explores the use of emojis in financial sentiment analysis, focusing on the social media platform StockTwits. Emojis, increasingly prevalent in digital communication, have potential as compact indicators of investor sentiment, which can be critical for predicting market trends. Our study examines whether emojis alone can serve as reliable proxies for financial sentiment and how they compare with traditional text-based analysis. We conduct a series of experiments using logistic regression and transformer models. We further analyze the performance, computational efficiency, and data requirements of emoji-based versus text-based sentiment classification. \rev{Using a balanced dataset of about 528{,}000 emoji-containing StockTwits posts, we find that emoji-only models achieve F$_1 \approx 0.75$, lower than text–emoji combined models (F$_1 \approx 0.88$) but with far lower computational cost---a useful feature where time is at premium, like in High Frequency Trading.} Furthermore, certain emojis and emoji pairs exhibit strong predictive power for market sentiments, demonstrating over 90\% accuracy in predicting bullish or bearish trends. Finally, our research also reveals extreme statistical differences in emoji usage between financial and general social media contexts, stressing the need for domain-specific sentiment analysis models.

\freefootnote{This submission is an extended version of the paper: 
``The Role of Emojis in Sentiment Analysis of Financial Microblogs'' presented at the 4\textsuperscript{th} IEEE International Conference on Intelligent Data Science Technologies and Applications, held in Kuwait in October 2023, and awarded with the Best Student Paper Award. \url{https://doi.org/10.1109/IDSTA58916.2023.10317863}
}
\end{abstract}





\maketitle

\section{Introduction}
\label{sec:intro}


Emojis have become a ubiquitous feature of human culture, acting as compact symbols that express a wide range of emotions, objects, ideas, and intentions in digital communication. More than 90\% of the online population uses emojis~\cite{UnicodeEmojiFrequency}. More than 20\% of Tweets contain emojis~\cite{BroniEmojiUse}. 
Billions of emojis are sent daily on Facebook Messenger~\cite{BurgeEmojis}. 
Emojis accompany around 40\% of text on Instagram, with their usage rapidly increasing~\cite{DimsonEmojineering}. 

This proliferation of emojis is not confined to casual communications alone; the realm of financial social media has been impacted by emoji use as well. 
Emojis, now integral to the lexicon of online communication, can serve as highly valuable indicators to analyze financial contexts. 
They offer rich information about emotions and opinions \rev{of} users, conveying investor sentiment and market perceptions~\cite{mahrous2023role,mahmoudi2022comprehensive}.
This understanding of market sentiment can be highly valuable for predicting or analyzing market trends and can thus be used in trading algorithms, financial analysis, or investment strategies~\cite{bashir2024investor,jun2024predicting}.
VanEck's Social Sentiment ETF\footnote{\url{https://www.vaneck.com/us/en/investments/social-sentiment-etf-buzz/overview/}}, a fund with tens of millions of dollars, is solely based on investor sentiment aggregated from online sources such as social media.
Financial social media sentiment indices are also strongly followed by market participants and offered by prestigious professional service firms such as Bloomberg and Thomson Reuters.

While emojis have been studied heavily in general contexts, they \rev{remain} understudied in the context of financial social media. Financial social media involves a specialized vocabulary of words and symbols that largely differs from general-purpose use, necessitating context-specific analysis. Moreover, unlike previous studies that merely enhance textual sentiment analysis using emojis as additional features, our study shifts the focus to investigating emojis as standalone indicators of financial sentiment. \rev{Treating} emojis as standalone indicators of financial sentiment is important for several reasons. If emojis alone can provide meaningful sentiment signals, they could serve as a more intuitive, faster, more efficient, and language-independent tool for financial analysis. Moreover, they could help analyze posts that do not include text but include only emojis.

Since emojis constitute a much smaller and simpler set of symbols compared to full text, processing them can be significantly faster. In fast-paced financial markets, where microseconds can make a difference in trading decisions, faster sentiment analysis can provide a significant competitive advantage~\cite{aquilina2022quantifying}. Working with emojis reduces the complexity of the data, leading to more efficient algorithmic processing. With fewer unique symbols than there are words, models can be trained and deployed with lower computational overhead. This efficiency not only speeds up analysis but also reduces the costs associated with large-scale data processing. Due to the lower complexity, sentiment analysis models based solely on emojis might be effectively trained using smaller datasets than those required for text-based models. This could be especially advantageous in specialized contexts where large volumes of labeled data are hard to obtain. Emojis can also transcend language barriers because they are universally recognized symbols~\cite{Kejriwal:2021:Biases, Guntuku:2019:Biases}. This means that sentiment analysis based solely on emojis can be applied globally without the need for translation or adaptation to different languages, making the analysis models more applicable in diverse international markets. Moreover, many social media posts, particularly on platforms where brevity is required, may consist mainly of emojis. By focusing on emojis as standalone sentiment indicators, analysts can capture the sentiment of these posts, which traditional text-based approaches would fail to analyze properly.\\


\noindent This leads us to the following research question.\\[-1em]

\noindent\textsl{What is the role of emojis in financial sentiment analysis, with respect to performance, computational efficiency, data requirements, patterns of use; and what are the differences between financial and generic use?}\\

\noindent Specifically, we seek to answer the following sub-questions.
\begin{itemize}
\item[Q1] How well can emojis \textsl{alone} serve as a proxy for public sentiment in finance?
\item[Q2] How does the computational speed of sentiment classification using only emojis compare to text-based methods? 
\item[Q3] How much data is needed to successfully train logistic regression and transformer-based models to predict financial sentiment trends (``bullish'', ``bearish'') in social media micro-messages using text-only, emoji-only, and text-with-emoji data?
\item[Q4] How significant is the difference in emoji usage between financial social media and non-financial social media?
\end{itemize}


\subsection{\rev{Contributions}}

Through addressing the mentioned research questions, we make the following contributions to the existing literature.



\begin{itemize}
    \item \textbf{Superior Financial Sentiment Classification Accuracy}. \rev{Using a balanced dataset of 528{,}000 StockTwits posts (evenly split between bullish and bearish), 
    our best text+emoji model achieves an F$_1$ score of 0.88, while the emoji-only variant reaches F$_1 \approx 0.75$. Although lower in accuracy, the emoji-only model retains most of the predictive power at a fraction of the computational and data demands,  demonstrating that emojis alone encode substantial sentiment information in financial communication.}     
    \item \textbf{Efficacy of Emoji-Only Models.} We demonstrate that emoji-only models can achieve accuracy comparable to text-based models but at substantially lower data requirements. As little as 1,000 posts used for training can be sufficient to achieve competitive accuracy, but at a significantly decreased training and inference time (cf.\ Figs.~\ref{fig:sampleSize} and~\ref{fig:sampleSizeTransformer}), especially if logistic regression is used. This gain in inference speed is highly relevant in contexts such as high-frequency trading.
    \item \textbf{Robust Emoji Sentiment Lexicon Construction.} 
    We employ a simple yet effective method to develop an emoji sentiment lexicon. In Section~\ref{sec: emoji_sentiment_scores}, we demonstrate that certain emojis and emoji pairs have strong predictive power (above 90\%) for sentiment classification. This facilitates rapid sentiment analysis.
    \item \textbf{Distinct Emoji Usage Patterns in Financial Contexts.} We identify extreme differences in emoji usage between financial and general social media contexts (cf.\ Section~\ref{sec:emoji_usage_StockTwits_versus_twitter}). Compared to previous research, our method related to this difference is more direct and explainable.
    \item \textbf{Analysis of Various Machine Learning Models.} We also evaluate various models, highlighting the efficiency of logistic regression compared to transformers in emoji analysis and addressing tokenization issues in Hugging Face models (cf.\ Subsection~\ref{subsec:Transformer}).

     \item \textbf{Data Sharing.}\rev{The code and data needed to replicate our results are publicly available at \url{https://github.com/AhmedMahrous00/finmoji_replication}}.
\end{itemize}

\subsection{\rev{Organization}}
The rest of this paper is organized as follows. Section~\ref{sec:related_work} presents existing research on emojis in sentiment analysis, particularly within the financial domain, highlighting key gaps that our paper addresses. Section~\ref{sec:data_and_methods} details our methodology and data, including data collection, pre-processing, and a section-by-section description of the methodologies employed. Section~\ref{sec:emoji_usage_descriptive_analysis} provides a descriptive analysis of emoji usage in our StockTwits data, detailing trends in emoji frequency, sentiment-based emoji distributions, and the comparative complexity of textual versus emoji data. This section motivates the value of incorporating emojis in financial sentiment analysis. Section~\ref{sec:emoji_usage_StockTwits_versus_twitter} compares emoji usage patterns between StockTwits and $\mathbb{X}$ (formerly Twitter)\footnote{In this paper, we use the terms ``Twitter'' and ``$\mathbb{X}$'' interchangeably to refer to the same platform.}, showcasing differences in emoji distributions and statistical tests. This section motivates the importance of developing models specific to financial platforms like StockTwits, rather than applying models designed for general social media platforms such as $\mathbb{X}$ (Twitter). To further illustrate the value of emojis in financial sentiment analysis, Section~\ref{sec: emoji_sentiment_scores} demonstrates single and pair emoji sentiment polarity by assessing their presence in bullish versus bearish posts. Section~\ref{sec:sentiment_analysis_algorithms} compares the accuracy and speed of various sentiment analysis algorithms applied to text and emoji data. Subsection~\ref{subsec:logistic} evaluates the performance of logistic regression models trained on text, emojis, and a combination of both. Subsection~\ref{subsec:Transformer} replicates the analysis for a transformer-based model. This subsection also explores the impact of training sample size and sequence length on model performance. Subsection~\ref{subsec:other_models} presents the performance of various machine learning models for emoji-based sentiment analysis, demonstrating the consistency and robustness of our findings across different models. Subsection~\ref{subsec:huggingfaceModelsComparison} compares the performance of various transformer-based sentiment analysis models on text and emoji data, demonstrating the rationale for selecting Twitter-RoBERTa as the primary model. Section~\ref{sec: discussion} synthesizes the key findings of this study, explores their implications, acknowledges limitations, and outlines potential directions for future research. Section~\ref{sec: conclusions} concludes.

\section{Related Work}
\label{sec:related_work}

Sentiment analysis of social media microblogs proved valuable across diverse domains, such as politics, marketing, and finance~\cite{du2024financial, demirel2025optimism, gunduz2025metaverse}. Researchers have employed both lexicon-based techniques (e.g., VADER, NRC, SO-CAL)~\cite{kahraman2025covid, kahraman2024exploring} and advanced deep learning models (e.g., BERT transformers)~\cite{gardazi2025bert} to interpret microblogs' sentiment. The advantage of lexicon-based techniques is that they are generally more interpretable and significantly faster/cheaper~\cite{van2025advantages}.

The value of emojis in sentiment analysis is increasingly gaining attention~\cite{zhou2024semantics}. While extensive research has explored the use of emojis in general contexts, their application in financial sentiment analysis remains less examined. This section synthesizes related research, highlights gaps, and reviews the studies that are most closely related to our work.

Research on the use of emojis in financial sentiment analysis is relatively scarce. We identified only approximately 40 publications that \textsl{mention} emojis in the context of finance. Most of this work uses emojis in addition to text to enhance sentiment analysis. Several studies~\cite{mahmoudi2018deep,omar2023uncover,renault2020sentiment}
demonstrate that the inclusion of emojis can significantly improve the accuracy of investor sentiment analysis.

Emojis have also been found useful in predictive systems for financial markets. For example, \citet{galvez2017assessing} analyze financial sentiment on an Argentinean stock message board, converting emojis to textual descriptions during preprocessing. They find that emojis are valuable features for predicting stock prices. Similarly, \citet{strych2022emojis} demonstrate that emojis, even when analyzed independently of text, possess standalone predictive value for stock prices based on their analysis of financial posts on Reddit. \citet{vamossy2021investor} uses emojis to help classify StockTwits posts into seven emotional states: neutral, happy, sad, anger, disgust, surprise, and fear. His analysis shows that these emotional states are predictive of company earnings and stock returns. For cryptocurrency markets, \citet{zuo2024emoji} utilize GPT-4 and BERT to analyze emoji sentiment on Twitter and demonstrate its predictive power for Bitcoin price movements and market volatility.

Furthermore, emojis contribute to the credibility assessment of financial social media accounts. \citet{bouadjenek2023user} discover distinct emoji usage patterns among StockTwits users who are more credible and give more accurate stock market predictions. \citet{shiri2023meme} extend this to Twitter, suggesting that emoji usage can be used to distinguish between genuine and spam posts on financial social media.

The inclusion of emojis into sentiment lexicons tailored for financial social media has been another topic of research. Chen et al.~\cite{chen2018ntusd, chen2019makes} have constructed a sentiment dictionary for StockTwits, incorporating 115 emojis, to capture the sentiment of financial social media. \citet{kulakowski2023sentiment} develop a ``language-universal cryptocurrency emoji sentiment lexicon'', abstracting from the natural language used in posts. 




Table~\ref{tab:literature_review} lists eight papers---filtered from the broader research work---that are most relevant to our research, as they utilize emojis in sentiment analysis algorithms for financial social media.

\begin{table}[htb]
\centering
\caption{\label{tab:literature_review}Review of Sentiment Analysis Studies Using Emojis in Financial Contexts. This table compares various sentiment analysis models across different studies, stating the models used, test data size, data sources, highest recorded accuracies, and whether they address sub-questions Q1 to Q4 (cf.\ Section~\ref{sec:intro}). Abbreviations: NB = Na\"{i}ve Bayes, Reg = Regression, SVM = Support Vector Machine, RF = Random Forest, MLP = Multilayer Perceptron, RNN = Recurrent Neural Network, CNN = Convolutional Neural Network, Tr = Transformer, Rule = Rule-based Algorithm. ``k'' stands for thousand; ``unbal/bal'' refers to unbalanced/balanced datasets; AC = accuracy; AUC = area under the curve; MCC = Matthews correlation coefficient; F$_1$ = F$_1$ Score. The test data size and performance are recorded for emoji-only analysis if the study includes such analysis and for text-with-emoji analysis otherwise. The highest achieved performance is recorded for studies with different models. \rev{Our models reach the highest reported performance (F$_1$ = 0.88) among all reviewed works.}}

\footnotesize{
    \setlength{\tabcolsep}{4pt}
    \resizebox{\textwidth}{!}{
    \begin{tabular}{cccccccccccc}
    \toprule
    \textbf{Ref.} & 
    \textbf{Models} & 
    \textbf{Source} & 
    \textbf{Test Data} & 
    \textbf{Performance} & 
    \textbf{Q1} & 
    \textbf{Q2} & 
    \textbf{Q3} & 
    \textbf{Q4} & 
    \textbf{EmojiOnly} & 
    \textbf{TextOnly} & 
    \textbf{Text+Emoji} \\
    \midrule
    \cite{renault2020sentiment} & 
    \makecell{NB, Reg\\ SVM, RF\\ MLP} & 
    StockTwits & 
    250k (bal.) & 
    \makecell{0.50 MCC} & 
    \ding{55} & 
    \ding{55} & 
    \ding{51} & 
    \ding{55} & 
    \ding{55} & 
    \ding{51} & 
    \ding{51} \\ 
    \hline
    
    \cite{mahmoudi2018deep} & 
    \makecell{NB, Reg,\\ SVM, RNN,\\ CNN} & 
    StockTwits & 
    \makecell{566k (unbal.)\\ 434k (bal.)} & 
    0.49 MCC & 
    \ding{55} & 
    \ding{55} & 
    \ding{55} & 
    \ding{51} & 
    \ding{55} & 
    \ding{51} & 
    \ding{51} \\
    \hline
    
    \cite{nasekin2020deep} & 
    RNN & 
    StockTwits & 
    564k (unbal.) & 
    \makecell{0.90 F$_1$ (bullish)\\ 0.58 F$_1$ (bearish)} & 
    \ding{55} & 
    \ding{55} & 
    \ding{55} & 
    \ding{55} & 
    \ding{55} & 
    \ding{55} & 
    \ding{51} \\
    \hline
    
    \cite{wilksch2023pyfin} & 
    \makecell{Tr, Reg,\\ SVM, Rule} & 
    \makecell{$\mathbb{X}$ (Twitter)\\ StockTwits} & 
    \makecell{2,500 \\ 2,510 } & 
    \makecell{0.83 AUC \\ 0.73 AUC} & 
    \ding{55} & 
    \ding{55} & 
    \ding{55} & 
    \ding{51} & 
    \ding{55} & 
    \ding{55} & 
    \ding{51} \\
    \hline
    
    \cite{chen2018ntusd} & 
    Rule & 
    StockTwits & 
    2,030 (unbal.) & 
    0.40 F1 & 
    \ding{55} & 
    \ding{55} & 
    \ding{55} & 
    \ding{51} & 
    \ding{55} & 
    \ding{55} & 
    \ding{51} \\
    \hline
    
    \cite{kulakowski2023sentiment} & 
    Tr, SVM & 
    StockTwits & 
    11,984 (unbal.) & 
    0.48 F1 & 
    \ding{51} & 
    \ding{55} & 
    \ding{55} & 
    \ding{55} & 
    \ding{51} & 
    \ding{51} & 
    \ding{55} \\
    \hline
    
    \cite{mahmoudi2022comprehensive} &
    \makecell{SVM, Reg,\\ RF} & 
    StockTwits & 
    Undeclared & 
    \makecell{0.47 F1} & 
    \ding{51} & 
    \ding{55} & 
    \ding{55} & 
    \ding{51} & 
    \ding{51} & 
    \ding{51} & 
    \ding{51} \\
    \hline
    
    {\bf Ours} & 
    Tr, Reg & 
    StockTwits & 
    528k (bal.) & 
    0.88 F1 & 
    \ding{51} & 
    \ding{51} & 
    \ding{51} & 
    \ding{51} & 
    \ding{51} & 
    \ding{51} & 
    \ding{51} \\
    \bottomrule
    \end{tabular}
}
}
\end{table}





\citet{renault2020sentiment} analyzes a balanced dataset of 1 million StockTwits posts, finding that including emojis, punctuation, and bigrams significantly enhances sentiment classification accuracy. Notably, his findings suggest that more complex and time-consuming methods, such as random forests or neural networks, do not necessarily outperform simpler approaches like multinomial Na\"{i}ve Bayes or logistic regression. Furthermore, he observes diminishing increases in sentiment classification accuracy with increasing data sizes. A critical aspect of Renault's research is his meticulous evaluation of different preprocessing techniques, revealing that standard tasks such as removing stopwords using the NLTK\footnote{\url{https://www.nltk.org}} stopwords corpus, Part-Of-Speech (POS) tagging, and stemming fail to improve or even negatively impact accuracy in financial social media \cite{renault2020sentiment}. 

Similarly, \citet{mahmoudi2018deep} utilize a vast StockTwits dataset, which exceeds 63 million posts, to develop a state-of-the-art sentiment classifier for financial social media. They report a significant improvement in accuracy, with a 44--47\% higher Matthews correlation coefficient (MCC)~\cite{Matthews} when emojis are included. Contrary to Renault's findings~\cite{renault2020sentiment}, they demonstrate that neural networks outperform simpler methods in financial sentiment analysis. Furthermore, they emphasize the importance of domain-specific models for financial social media sentiment analysis, showing that domain-specific word embeddings significantly outperform general-domain ones.

\citet{wilksch2023pyfin} utilize a much smaller dataset of 10,000 financial posts on Twitter, which are manually labeled by the authors, to experiment with different sentiment classification models and to develop their own. They affirm that domain-specific models perform significantly better than ``generic off-the-shelf models
like VADER~\cite{VADER},'' thus reinforcing the importance of tailored approaches in financial sentiment analysis. In a similar observation to that of \citet{renault2020sentiment}, they demonstrate that simpler models, such as logistic regression, can match the accuracy of more complex models like deep neural networks, but with greater computational efficiency and reduced risks of overfitting. However, they also note significant limitations with logistic regression, particularly its shortcomings in learning more complex word dependencies, such as negation, which could impair its performance in analyzing certain posts. Furthermore, their research reveals that models trained on data from financial Twitter posts underperform when applied to StockTwits data, illustrating the challenge of platform dependency in sentiment analysis.


\citet{nasekin2020deep} also explore developing a sentiment classifier for financial social media, utilizing around 560,000 labeled StockTwits posts. They incorporate emojis in their model and manage to achieve their highest F$_1$ score of 0.90 using an LSTM~\cite{LSTM,Gers:2000:LTF} model, but their model is heavily skewed towards bullish \rev{sentiment}, achieving a low F$_1$ score of only 0.58 on bearish posts. \citet{chen2018ntusd} utilize 334,000 labeled StockTwits posts to construct a financial microblogging sentiment dictionary. Their dictionary includes more than 100 emojis and performs better than the seminal dictionary developed on formal financial texts by \citet{loughran2011liability}.

While the previously-mentioned studies incorporate emojis alongside text in their models, the studies by \citet{kulakowski2023sentiment} and \citet{mahmoudi2022comprehensive} uniquely stand out within our review. They are the only ones who test financial sentiment analysis using emojis exclusively, without any accompanying text.

\citet{kulakowski2023sentiment} fine-tune the transformer-based BERTweet~\cite{bertweet} model utilizing 1.875 million posts (labeled and unlabeled) from StockTwits. They also develop a language-universal cryptocurrency emoji (LUKE) sentiment lexicon using 91,758 StockTwits posts. They tested sentiment analysis using only this emoji lexicon for sentiment analysis, but achieved a low F$_1$ score of 0.4816.

\citet{mahmoudi2022comprehensive} utilize more than 6 million StockTwits posts (each containing at least one emoji) to perform a comprehensive study of emojis in financial social media. They show that using domain-specific (in contrast to domain-independent) emoji embeddings is highly valuable for both sentiment classification and price predictions. They also demonstrate the significant difference between financial and general emojis using cluster analysis. They develop an emoji sentiment lexicon and use emoji-only data to test their sentiment analysis algorithm. However, they only achieve relatively low accuracy, with an F$_1$ score of 0.39.

While these works show that emojis are useful for sentiment analysis, significant gaps and limitations still exist within this area of study. Firstly, there is a noticeable inconsistency in the methodologies used across studies. For instance, the sentiment analysis problem is converted to binary classification in some studies and non-binary in others. Different studies also employ various balancing techniques, choosing to oversample, undersample, or analyze unbalanced data. Furthermore, the accuracy metrics reported are often inconsistent, hindering direct comparisons. The wide variety of model specifications and pre-processing options contributes to this inconsistency. With respect to emojis, some studies convert them to their textual descriptions, while others treat them as independent symbols. Additionally, some studies use domain-general models such as VADER~\cite{VADER}, which have been shown to be less effective in financial sentiment analysis. Many of the studies also rely on relatively small and unbalanced datasets, which limits the robustness and generalizability of their findings. The lack of standardization and oversight of critical details in methodologies contributes to varying sentiment classification accuracies, with some papers achieving very low accuracy.

Among the many gaps existent around this topic, the ones focused on in our analysis (and listed in our research sub-questions), remain insufficiently addressed. While \citet{kulakowski2023sentiment} and \citet{mahmoudi2022comprehensive} have addressed sub-question Q1, they have achieved very low accuracy in their sentiment classification. \citet{mahmoudi2022comprehensive} also address this gap very briefly as a side-analysis of their other main ideas and, therefore, ignore mentioning many details. \citet{renault2020sentiment} addresse\rev{s} sub-question Q3 only partially, not testing the dataset size requirements of using emojis-only or text-only data. \citet{renault2020sentiment} also does not mention whether the test set was kept fixed for all the dataset sizes, a crucial detail of such an experiment. Sub-question Q4 has been addressed more extensively than the other sub-questions. However, most of the papers addressing it compare models using combined text and emoji data. Only \citet{mahmoudi2022comprehensive} focus exclusively on emojis, using a different methodology than ours. We believe that our methodology, presented in  Section~\ref{sec:emoji_usage_StockTwits_versus_twitter}, can help demonstrate and quantify the difference in emoji usage between domains in a more explainable manner. To the best of our knowledge, no study other than ours has addressed sub-question Q2.



\subsection*{\rev{\bf Summary of Novelty.}}
\rev{
In summary, our main contributions are demonstrating the \emph{effectiveness and efficiency} of emoji-only sentiment analysis models in the context of financial social media. Most prior finance sentiment studies augment text with emojis but do not investigate an emoji-only model (e.g.,~\citet{renault2020sentiment}). In contrast, we evaluate emoji-only, text-only, and text+emoji models. As summarized in Table~\ref{tab:literature_review}, our models achieve a markedly higher overall performance (F$_1$ = 0.88) than all prior studies, most of which report F$_1$ scores or equivalent metrics below 0.50 on data sources similar ours. To the best of our knowledge, only \citet{mahmoudi2022comprehensive} and \citet{kulakowski2023sentiment} attempt emoji-only analysis, both being lexicon-based and both achieving F$_1$~$<0.5$. Relative to them, we show substantially stronger emoji-only effectiveness with F$_1$~$\approx0.75$. Notably, we find that a simple Logistic Regression model performs on par with a Transformer model in the emoji-only setting (F$_1{=}0.76$ in both cases), highlighting that emoji sentiment signals are linearly separable and do not require deep contextual embeddings.\\ 
In addition, we introduce formal cross-domain tests to compare between financial and non-financial social media, motivating finance-specific analysis---something prior studies did not do. We also provide practical insights, including a tokenizer audit across BERT-family models showing that most mistreat emojis, while Twitter-RoBERTa preserves them; such guidance can inform future sentiment-analysis research. While the previous two papers intuitively argue for the efficiency of using emojis, they do not empirically demonstrate it. We do, providing an integrated efficiency study: computational efficiency measurements (training/inference time), learning curves showing reduced data needs---an important advantage in data-scarce contexts---and context-length sensitivity demonstrating that emoji-only peaks at short contexts, further strengthening the efficiency argument. Taken together, our results demonstrate the effectiveness and efficiency of using emoji-only models in financial social media sentiment analysis in a way that previous studies did not.
}

\section{Data and Methods}
\label{sec:data_and_methods}

In this section, we describe our data and discuss the methodologies used in the different sections of this paper.


\subsection{Data}

We begin by gathering 18.5 million posts that were published by users on the social media platform StockTwits. Each one of these posts had the cash tag of at least one stock or cryptocurrency. As detailed in Table~\ref{tab:cashtags}, the composition of our sample mirrors the platform's user interests, with a significant portion dedicated to discussions surrounding the S\&P 500 index, major cryptocurrencies like Bitcoin, and popular meme stocks and coins such as GameStop and Dogecoin. The remaining posts encompass conversations about large-cap stocks like Apple, established cryptocurrencies like Ethereum, and indices tracking assets like gold.

\begin{table}[ht]
\centering
\caption{\label{tab:cashtags}Distribution of Posts by Financial Asset. This table illustrates the percentage of posts within our dataset mentioning particular financial assets.}
\begin{tabular}{lccccccc}
\toprule
& S\&P 500 & Bitcoin & Dogecoin & Apple & GameStop & Shiba Inu & Others \\
\midrule
Percentage & 33\% & 15\% & 12\% & 10\% & 7\% & 7\% & 16\% \\
\bottomrule
\end{tabular}
\end{table}

\begin{figure}[ht]
   \centering
   \includegraphics[width=\linewidth]{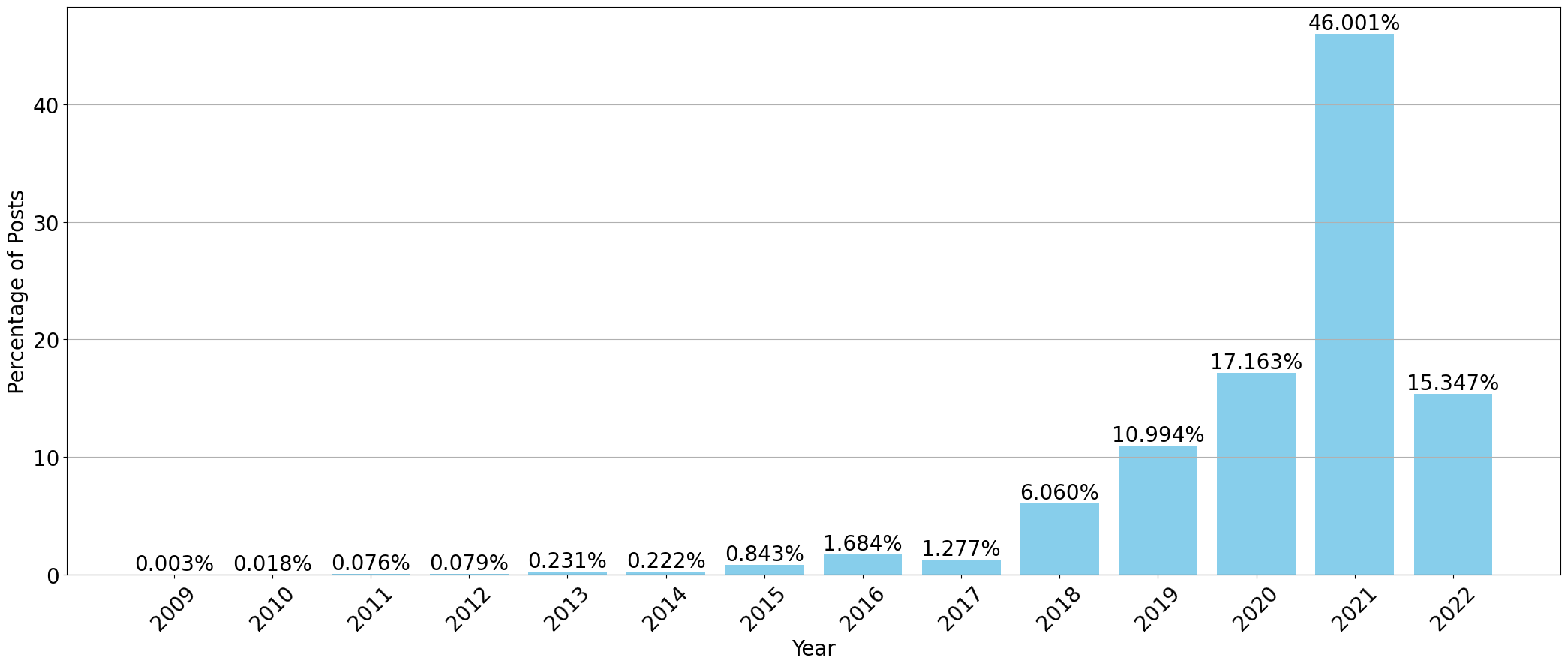}
   \caption{Distribution of Posts by Year of Publishing. This figure shows the percentage of total posts per year from 2009 to 2022 within our dataset.}
    \label{fig:PostsByYear}
\end{figure}

Spanning from February 2009 to October 2022, our dataset provides a comprehensive view of financial social media activities over time. Fig.~\ref{fig:PostsByYear} illustrates the distribution of posts by year, highlighting the surge in activity from 2018 onwards, with 2021 accounting for nearly half of the total posts. We attribute this to the COVID-19 pandemic, with lockdowns and home offices driving the population of many countries into a more online-centric lifestyle, paired with shop closures making spending money in physical outlets harder or impossible.

\begin{figure}[ht]
   \centering
   \includegraphics[width=\linewidth]{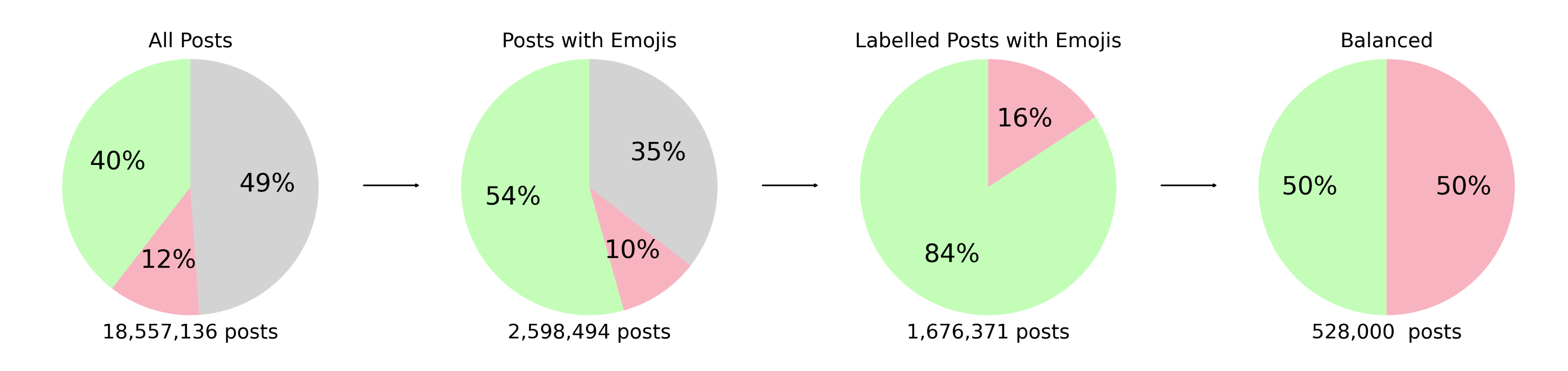}
   \caption{Data Filtration Process. This series of pie charts illustrates the progressive stages of data filtering from the initial dataset to the final balanced sample. Colors represent sentiment classification: green for bullish, red for bearish, and grey for unlabeled posts.}
    \label{fig:datafilteration}
\end{figure}

While our initial dataset comprises 18.5 million posts, only 14\% incorporate emojis, narrowing our focus to approximately 2.5 million posts. Further refinement was necessary as not all posts included user-labeled sentiment, a crucial element for supervised machine learning. This results in a subset of 1.7 million labeled posts. However, a significant imbalance is present, with bullish sentiment dominating at 85\%. To some extent, this aligns with the global economic situation of that time---The S\&P 500 doubled from 2018 to 2021, gaining 28.7\% in 2021 alone. However, such an imbalance, whether rooted in economic reasons or not, hinders the generalization of a model, especially when market trends change. Thus, to ensure the effectiveness of our sentiment analysis algorithms, we opted to balance the data by undersampling the majority class. Consequently, our final dataset contains 528,000 posts evenly split between bullish and bearish sentiment, each containing at least one emoji and referencing a specific stock or cryptocurrency via hashtag. Fig.~\ref{fig:datafilteration} illustrates our process.


Several factors motivated our choice of StockTwits as the data source. Primarily, StockTwits offers a unique advantage in that it allows users to explicitly label their posts with their sentiment---either bullish, indicating an expectation of price increases, or bearish, suggesting anticipated price declines. This readily available sentiment data proves invaluable for training supervised learning algorithms. Furthermore, research has demonstrated that StockTwits data exhibits a lower prevalence of low-quality content, such as spam, compared to platforms like Twitter, as evidenced by \citet{bouadjenek2023user}.

\begin{figure}[ht]
   \centering
   \includegraphics[width=\linewidth]{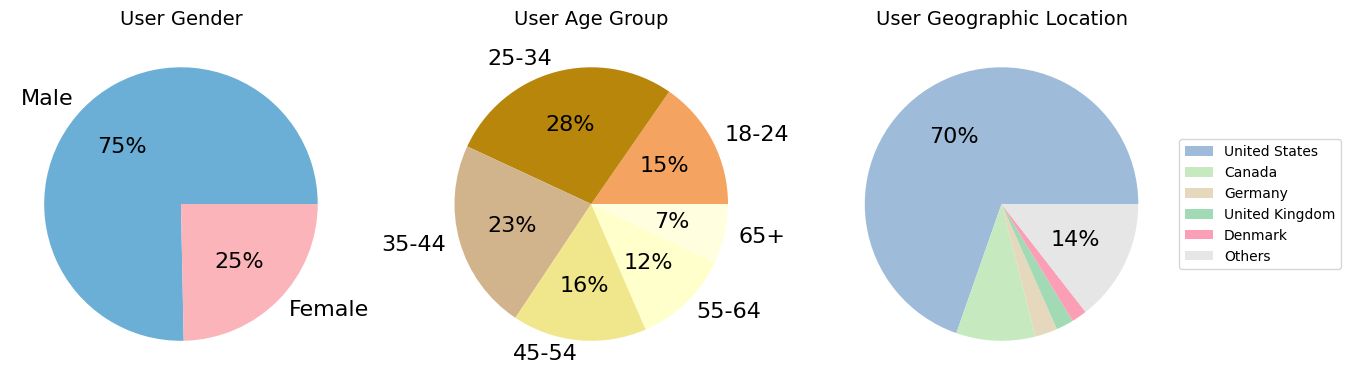}
   \caption{Demographic Breakdown of StockTwits Users. This figure illustrates the demographic composition of StockTwits users, highlighting gender distribution, age groups, and geographic locations. Data sourced from \cite{Similarweb}.}
    \label{fig:StockTwitsDemographics}
\end{figure}

Fig.~\ref{fig:StockTwitsDemographics} provides a visual representation of StockTwits user demographics based on data from web analytics company Similarweb~\cite{Similarweb}. Notably, the platform exhibits a predominantly male user base. The age distribution is relatively diverse, with the largest segment comprising adults aged 25 to 44. Users primarily hail from English-speaking countries, particularly the USA, which aligns with the overwhelming majority of posts being in English. This demographic information is pertinent to our analysis as research has shown that informal language and emoji usage on social media can vary significantly across genders, age groups, and cultures~\cite{Kejriwal:2021:Biases, chen2024individual,Chen:2018:Gender}.

\subsection{Methodologies}
\subsubsection{Emoji Usage: Descriptive Analysis}

Section~\ref{sec:emoji_usage_descriptive_analysis} investigates emoji usage patterns within our StockTwits dataset. We analyze the distribution of emojis per post, considering both unique and total occurrences, as depicted in Fig.~\ref{fig:descriptiveStats}. This figure further illustrates the average monthly emoji usage from February 2009 to July 2022, encompassing all 18.5 million posts. To visualize emoji prevalence, we generate emoji clouds (Fig.~\ref{fig:clouds}) using the wordcloud package\footnote{\url{https://pypi.org/project/wordcloud/}} and the Symbola font for the emojis, focusing on the 2.6 million posts containing emojis. Lastly, Table~\ref{table:tokenlengthsPercentile} presents percentile distributions for the number of textual characters and emojis based on the balanced subset of 528,000 emoji-containing posts.

\subsubsection{Emoji Usage: StockTwits versus $\mathbb{X}$ (Twitter)}

Section~\ref{sec:emoji_usage_StockTwits_versus_twitter} compares emoji usage between StockTwits and $\mathbb{X}$ (Twitter). We analyze our StockTwits dataset alongside a publicly available Twitter dataset\footnote{\url{https://www.kaggle.com/datasets/rexhaif/emojifydata-en?resource=download\&select=emojitweets-01-04-2018.txt}}, employing the Mann-Whitney U~\cite{Mann:1947:U} and Kolmogorov-Smirnov~\cite{berger2014kolmogorov,Smirnov:1948} tests to assess differences in the relative frequency distributions of emojis. \rev{To test distributional differences in emoji use across platforms, we also compute the chi-square statistic and its effect size via Cram\'{e}r's $V$~\cite{Cramer:1946:V}: 
\begin{align}
V&= \sqrt{\chi^2 / \left(n \times \min\left(k - 1, r - 1\right)\right)},
\end{align} 
where $n$ is the total observation count and $k$ and $r$ are the numbers of columns and rows. In our case, columns and rows correspond to platform and emoji occurrences. This approach is appropriate because both variables are categorical; the chi-square test assesses whether their distributions differ beyond random variation, while Cram\'{e}r's~$V$ quantifies the strength of this association on a standardized $[0,1]$ scale.} \rev{To compare datasets of equal sample sizes, we randomly sampled 2.6 million tweets from the larger Twitter dataset. 2.6 million is also the number of StockTwits posts in our dataset containing at least one emoji.} The tests focus on the presence of emojis within posts, disregarding the number of repetitions. For these tests, our analysis considers the presence of an emoji within a post rather than its frequency, meaning that multiple occurrences of the same emoji within a single post are counted only once. 

\subsubsection{Emoji Sentiment Scores}
Section~\ref{sec: emoji_sentiment_scores} focuses on assessing the sentiment associated with individual emojis. We analyze the 528,000 user-labeled posts containing emojis, focusing on the 50 most frequently used emojis while disregarding multiple occurrences within a single post. For each of these emojis, we calculate a bullish and bearish sentiment score using the following formula.
\begin{align}
\text{Bullish (Bearish) Sentiment Score} &= 
\frac{\text{Number of bullish (bearish) posts containing the emoji}}
{\text{Total number of posts containing the emoji}} \nonumber
\end{align}
This approach provides a simple yet robust measure of the sentiment conveyed by each emoji.

Focusing on the top 50 most frequently used emojis enhances the reliability of our analysis, as less frequent emojis exhibit greater volatility in their sentiment scores. This selection strategy ensures a more robust assessment, with the least frequent emoji in our top 50 (\emoji{victory-hand}) still appearing in 3,242 posts.

We extend this sentiment polarity analysis to pairs of emojis as well. We find the 50 most frequently used emoji pairs (two emojis that are different but co-exist in the same post). We avoid double-or-more counting: if two emojis are mentioned multiple times in the same post, regardless of order, this is counted as only one co-occurrence. As an example, (\emoji{rocket} \emoji{gem} \emoji{gem} \emoji{rocket}) would be considered as only one co-occurrence of the pair (\emoji{gem} \emoji{rocket}). We use the same formula used for the single emoji analysis to calculate bullish and bearish sentiment scores for each pair of emojis. Again, using only the 50 most used emoji pairs increases the robustness of our analysis. The least utilized emoji pair from the top 50 (\emoji{skull} \emoji{fire}) is used in 1,154 different posts.

In Fig.~\ref{fig:numEmojis}, we calculate the percentage of bullish and bearish labels given to posts, grouped by the number of unique emojis they contain. For instance, the first group shows the posts that contain one unique emoji or repetitions of it; for this group and each of the other groups, we calculate the percentage of posts that are labeled as bullish and those that are labeled as bearish to construct the two bars. We do this for ten groups, with the first group having only one unique emoji and the last group having ten or more unique emojis.

\subsubsection{Emoji Sentiment Analysis Algorithms}
\begin{figure}[ht]
    \begin{center}
        \includegraphics[width=0.99\textwidth]{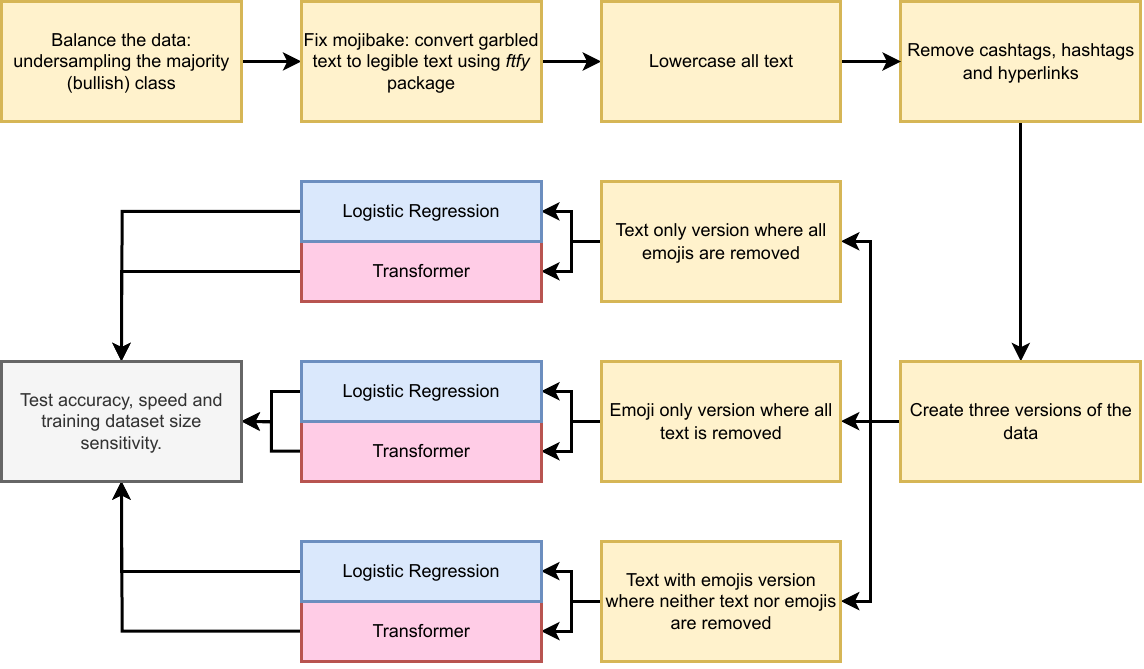}
    \end{center}
    \caption{Methodology at a Glance. Pre-processing (yellow) generates three datasets containing only text, only emojis, and text with emojis. We then feed each dataset into both a logistic regression model (blue, cf.\ Subsection~\ref{subsec:logistic}) and a transformer-based model (pink, cf.\ Subsection~\ref{subsec:Transformer}).}
    \label{fig:methodology}
\end{figure}

\subsubsubsection{Logistic Regression}
Subsection~\ref{subsec:logistic} uses the 528,000 labeled posts that include emojis. A regular expression tokenizer with a specific pattern (\verb/\w+|[^\s]/) is used to tokenize the data. This custom tokenizer is designed to capture word characters and any non-whitespace character, including emojis, separating them into individual tokens. To transform the tokenized data into a TF-IDF (Term Frequency–Inverse Document Frequency) matrix~\cite{TFIDF:1972}, we employ the TfidfVectorizer implementation available in Scikit-learn\footnote{\url{https://scikit-learn.org}}. This step transforms textual/emoji data into numerical form that can be processed by the logistic regression model. TF-IDF quantifies the importance of a token in a post relative to the entire collection of 528,000 posts. We use the following log-normalized formula to assign a number to each token.
\begin{align}
\text{TF-IDF}(\text{token}) &= 
\left( \frac{\text{Number of times the token appears in a post}}{\text{Total number of tokens in a post}} \right) \nonumber \\
&\quad \times  \log\left( 1+ \frac{\text{Total number of posts}}{\text{Number of posts containing the token}} \right)
\end{align}

We employ a standard train-test split, with 80\% of the data allocated for training and the remaining 20\% for testing using Scikit-learn's \texttt{train\_test\_split} function. We then train a logistic regression model on the training set for the purpose of binary classification, predicting whether a post expresses bullish or bearish sentiment. We keep the default decision threshold of 0.5.
We evaluate the performance of our model by comparing predicted and actual labels on the test set, utilizing metrics such as accuracy, precision, recall, and F$_1$ score derived from Scikit-learn's \texttt{classification\_report} and \texttt{confusion\_matrix} functions.

Table~\ref{logisticPerformance} additionally presents speed metrics for the logistic regression model, including both training and inference times. Training time refers to the duration required to train the model on a subset of 80,000 posts, while inference time measures the time taken to predict the sentiment of 20,000 posts. 

In Fig.~\ref{fig:sampleSize}, we present the accuracy of a logistic regression model for binary classification, which is trained using varying sample sizes. This model adheres to the methodologies described in the preceding sections. We test the model on a consistent and fixed dataset comprising 52,800 posts. To assess how much data is required to achieve sufficient accuracy, we randomly selected training sets of size 100, 1,000, 10,000, 100,000, and 400,000 posts from the original dataset

The aforementioned experiments were conducted on three variants of the data: posts containing only text (excluding emojis), posts consisting solely of emojis (excluding text), and posts that included both text and emojis.

\rev{To assess uncertainty in accuracy and speed metrics, we use a non-parametric bootstrap method: we draw $n$ samples with replacement from the test set, compute each metric, repeat $1000$ times, and report the 2.5$^{\text{th}}$ and 97.5$^{\text{th}}$ percentiles as 95\% confidence intervals.}

\subsubsubsection{Transformers}
For Subsection~\ref{subsec:Transformer}, we use 528,000 labeled posts that include emojis, with 10\% (52,800 posts) reserved for testing. The remaining 90\% are split into training (81\% of the original set) and validation sets (9\% of the original set). We employ the \texttt{cardiffnlp/twitter-roberta-base-sentiment-latest} model~\cite{barbieri2020tweeteval} and its associated tokenizer. This model, selected based on its superior performance on our dataset (cf.\ Table~\ref{table:huggingfacemodels}), builds on the RoBERTa-base architecture~\cite{RoBERTa} and is further pre-trained on approximately 124 million tweets and adapted for sentiment analysis. Since the model is initially designed for three-label sentiment classification, we adapt it for binary classification using the \texttt{RobertaForSequenceClassification} class from the Hugging Face\footnote{\url{https://huggingface.co}} transformers library. BERT-based models are chosen for our analysis due to their demonstrated state-of-the-art performance in sentiment analysis, as supported by previous research in both general social media \cite{khan2025sentiment, bashiri2024comprehensive} and financial contexts \cite{mahendran2025comparative, cicekyurt2025enhancing, du2024financial}.

Prior to training, each post is truncated or padded to 120 tokens. \rev{Based on the percentile distribution in Table~\ref{table:tokenlengthsPercentile}, 99\% of StockTwits posts in our dataset are under 484 characters ($\approx$ 100 BERT tokens), indicating that a sequence limit of 120 tokens almost guarantees full coverage. As shown in Table~\ref{tab:max-length}, extending context length beyond 50 tokens provides little F$_1$ accuracy improvement while significantly increasing computational demands. We therefore adopted 120 tokens as a conservative upper bound to ensure robustness across text-only and emoji-only settings. This value is also consistent with the 128-token window commonly used in large-scale tweet transformer models such as BERTweet~\cite{bertweet}.}

The model is then fine-tuned on our sample data for three epochs using the Hugging Face Trainer API, with a batch size of 32. The learning
rate is dynamically adjusted using a cosine scheduler with restarts, and a warm-up phase of 50 steps is included to gradually increase the learning rate at the beginning of training. We utilize mixed-precision training (FP16) and gradient accumulation to enhance computational efficiency. We evaluate the performance of our model against the validation dataset at 10 predefined checkpoints, with the best-performing model weights saved for final assessment on the test dataset using accuracy, precision, recall, and F$_1$ score.

For the speed metrics in Table~\ref{transformerPerformance}, we randomly selected 80,000 posts to measure training speed and 20,000 for inference speed. The same model configuration and tokenizer are used, with all posts truncated or padded to 120 tokens. We report training speed for a single epoch. 

\begin{figure}[ht]
   \centering
   \includegraphics[width=\linewidth]{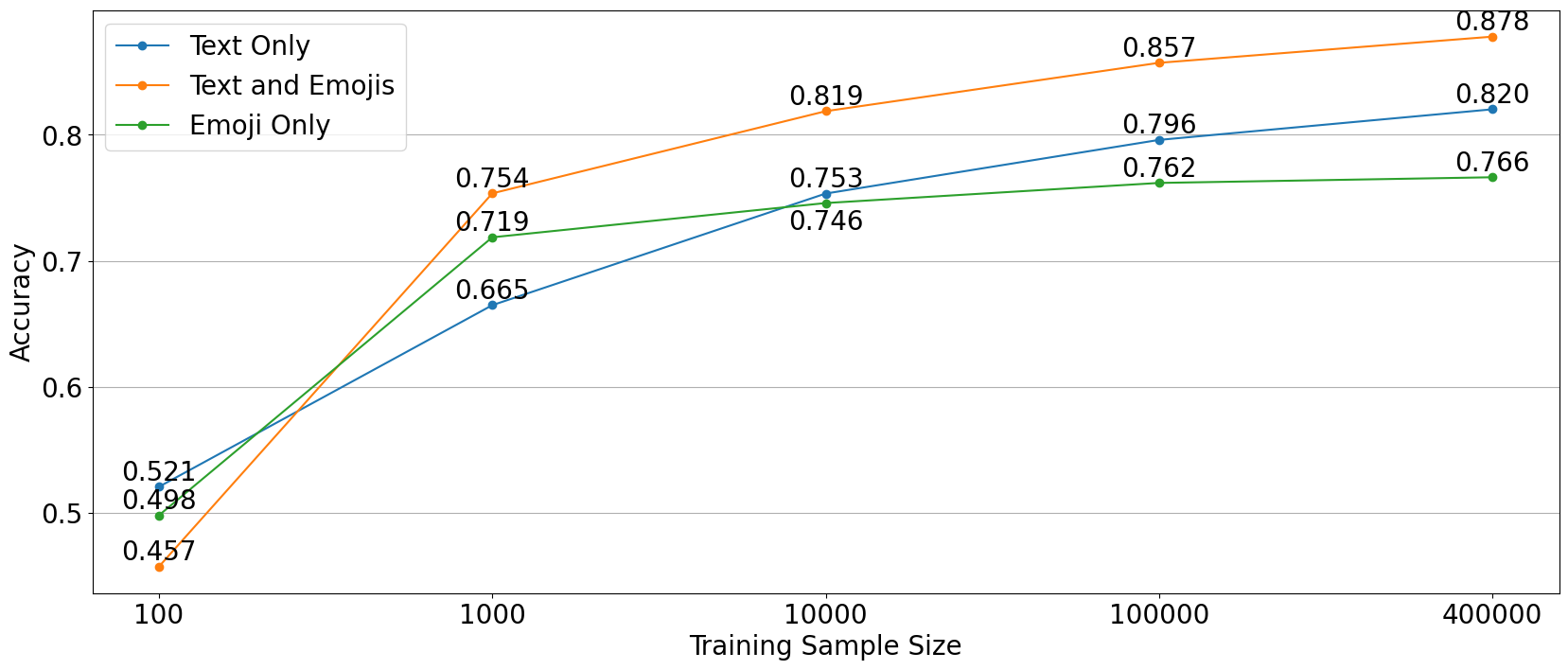}
   \caption{Impact of Training Sample Size on Transformer-based Twitter-RoBERTa Model Accuracy. This figure illustrates the relationship between training sample size and model accuracy for Twitter-RoBERTa models trained on text only (blue line), emojis only (green line), and text with emojis (orange line). Accuracy is evaluated on the same separate test set of 50,000 posts.}
   \label{fig:sampleSizeTransformer}
\end{figure}

In Fig.~\ref{fig:sampleSizeTransformer}, the identical model setup is applied, but we train on random subsets of the original training dataset: 100, 1,000, 10,000, 100,000, and 400,000 posts. Each subset is trained over three epochs, with accuracy measured on the original test dataset of 52,800 posts, and results are plotted in Fig.~\ref{fig:sampleSizeTransformer}.

All experiments were conducted on three data variants: posts containing only text (excluding emojis), posts consisting solely of emojis (excluding text), and posts including both text and emojis.

\rev{We use the same non-parametric bootstrap method as before to assess uncertainty, that is, we draw $n$ samples with replacement from the test set, compute each metric, repeat $1000$ times, and report the 2.5$^{\text{th}}$ and 97.5$^{\text{th}}$ percentiles as 95\% confidence intervals.}

\subsubsubsection{Other Tested Models} 
In Subsection~\ref{subsec:other_models}, we explore the performance of various machine learning algorithms on sentiment analysis using only emoji data. This analysis uses the same data variant that includes only emojis, as discussed in previous sections. The methodology mirrors that of the logistic regression model described in Subsection~\ref{subsec:logistic}, employing the same RegexpTokenizer (\verb/\w+|[^\s]/) and TF-IDF vectorizer. We utilize the dataset of 528,000 posts containing only emojis, with 80\% allocated for training and 20\% for evaluation.

We examine four different machine learning algorithms: Multinomial Na\"{i}ve Bayes (MNB), Random Forest Classifier (RF), LightGBM Classifier (LGBM)~\cite{LGBM}, and Linear Support Vector-Machine Classifier (LinearSVC). Each model is trained and evaluated using the same datasets as those used for the logistic regression model in Subsection~\ref{subsec:logistic}. Speed metrics are calculated similarly to those for the logistic regression models in Subsection~\ref{subsec:logistic} and the transformer models in Subsection~\ref{subsec:Transformer}, facilitating direct comparison. All experiments were conducted on the same hardware. The configurations for these models are set to their default parameters as specified in the scikit-learn module, except for the n\_jobs=-1 setting for the RF model to use all CPU cores during training.

\subsubsection{Huggingface Model Comparisons}

In Subsection~\ref{subsec:huggingfaceModelsComparison}, we employ the same methodology outlined in Subsection~\ref{subsec:Transformer}, using the same pipeline and the Hugging Face Trainer API but with a variety of different models available on the Hugging Face platform. For each model (\texttt{*}), we use the corresponding \texttt{*Tokenizer} and \texttt{*ForSequenceClassification} classes from the Hugging Face transformers library. A key distinction, however, is that for all models, except the one used in Subsection~\ref{subsec:Transformer}, the attached tokenizer excludes emojis by default. Consequently, we preprocess by ``demojizing'' emojis, which involves converting emoji symbols into their respective textual descriptions or aliases. This preprocessing step is executed using the \texttt{demojize} function from the Python \texttt{emoji} library\footnote{\url{https://pypi.org/project/emoji/}}. All models are then fine-tuned using the same datasets consisting of text only, emojis only, and both text and emojis.




\section{Emoji Usage: Descriptive 
Analysis}
\label{sec:emoji_usage_descriptive_analysis}

\begin{figure}[ht]
    \centering
    \subfloat[]{
        \includegraphics[width=0.48\textwidth]{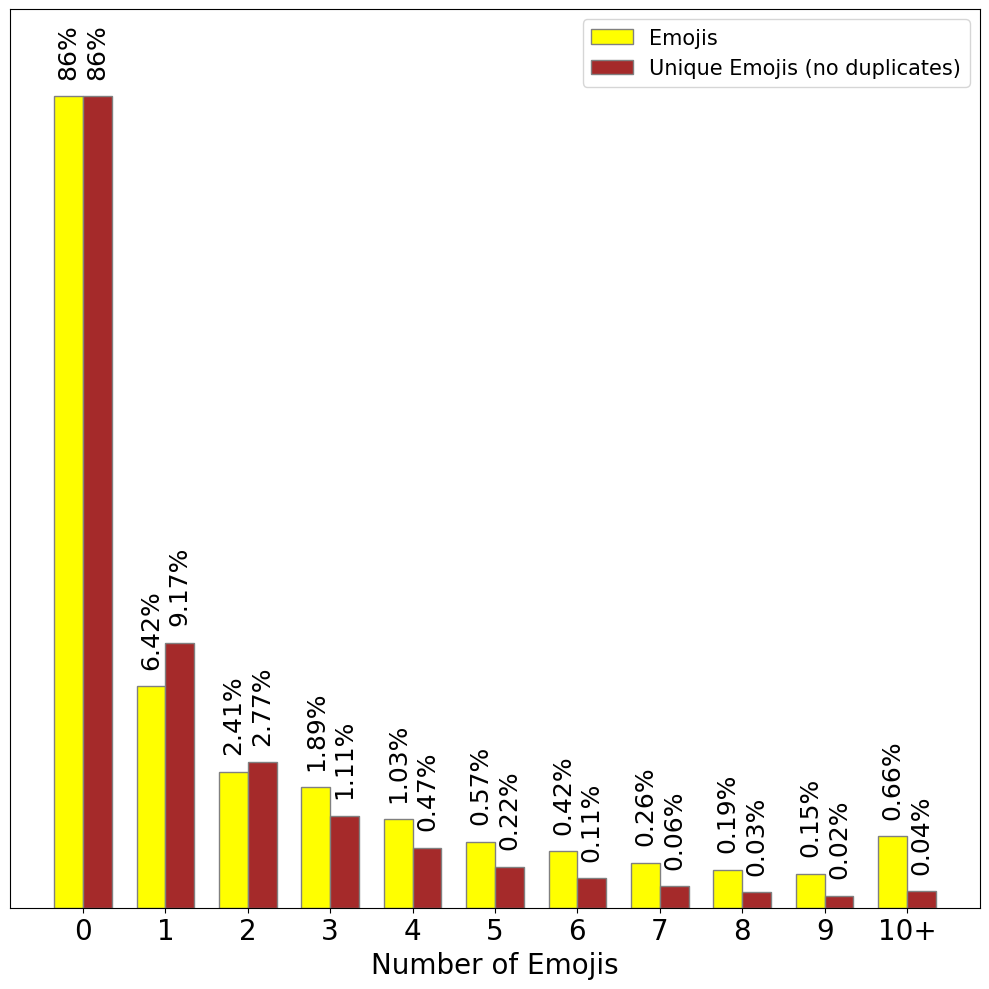}
    }
    \subfloat[]{
        \includegraphics[width=0.48\textwidth]{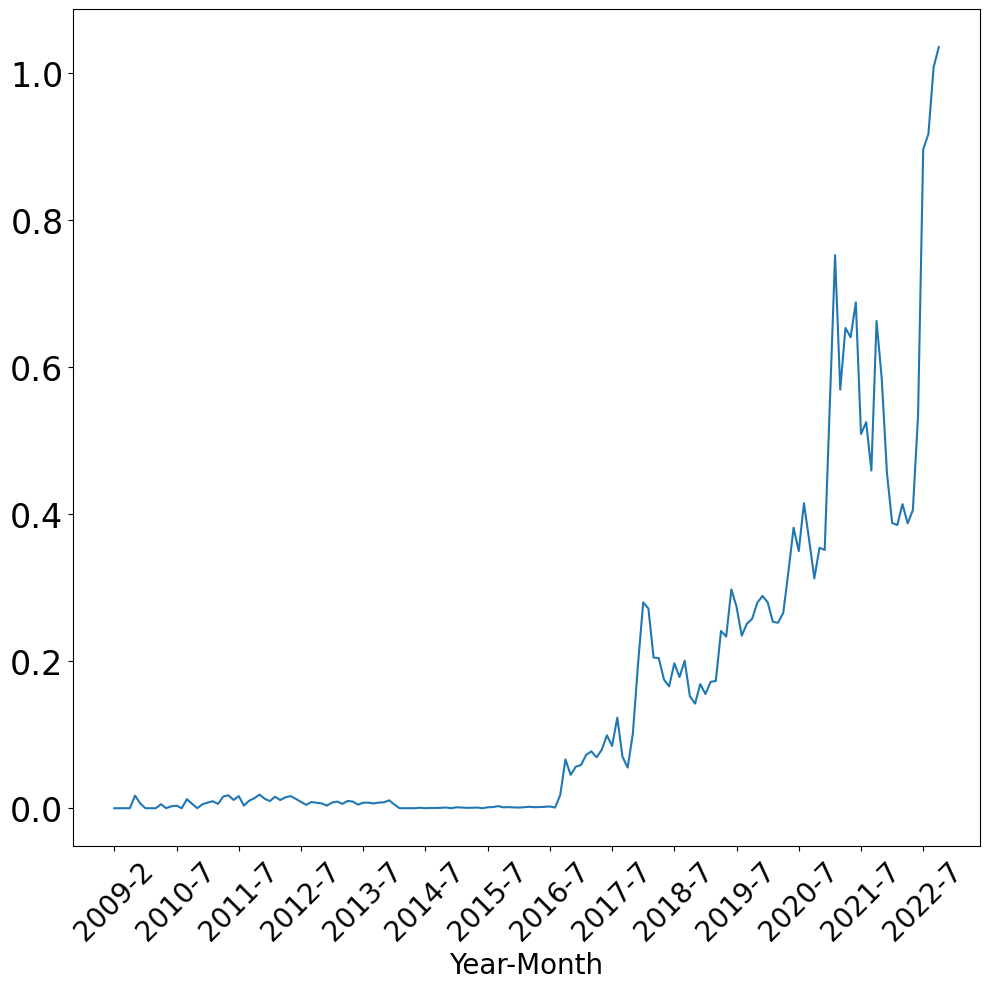}
    }
    \caption{Emoji Usage in our Dataset. (a) Bar charts illustrating the percentage of posts containing various numbers of emojis, comparing total emoji counts (yellow) and unique emojis (red). (b) A time series detailing the average number of emojis used per post over time.}
    \label{fig:descriptiveStats}
\end{figure}

In this section, we conduct a descriptive analysis of emoji usage in StockTwits posts. Approximately 15\% of the 18.5 million posts in our dataset include at least one emoji. The distribution of emojis per post shows a negative trend: the frequency of posts containing a given number of emojis decreases as the emoji count increases. This pattern is evident in Fig.~\ref{fig:descriptiveStats} (left) and applies whether considering the total number of emojis or the count of distinct emojis per post. Notably, a significant portion of posts contain more than 10 emojis, suggesting potential advertisement or spam activity. \citet{shiri2023meme} find that patterns of emoji usage help in detecting spam in financial tweets. Fig.~\ref{fig:descriptiveStats} (right) shows an explosive increase in the usage of emojis over time. From 2009 to 2016, the average number of emojis per post remained near zero. Subsequently, there was an exponential increase, with the average exceeding 1.0 by mid-2022. The rise in the usage of emojis is a well-documented phenomenon~\cite{BroniEmojiTime}, possibly caused by socio-cultural and technological shifts. The short-term variability in emoji usage over time presents an interesting area for further investigation.

\begin{figure}[ht]
    \centering
    \subfloat[Bullish]{
        \includegraphics[width=0.33\textwidth]{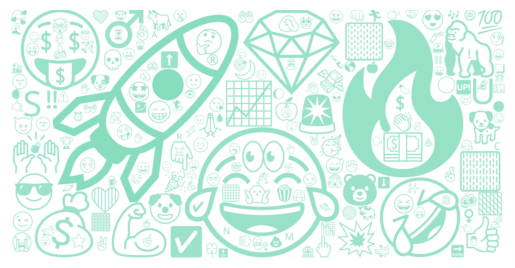}
    }
    \subfloat[Bearish]{
        \includegraphics[width=0.33\textwidth]{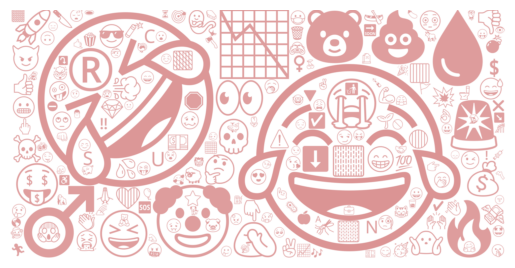}
    }
    \subfloat[Unlabeled]{
        \includegraphics[width=0.33\textwidth]{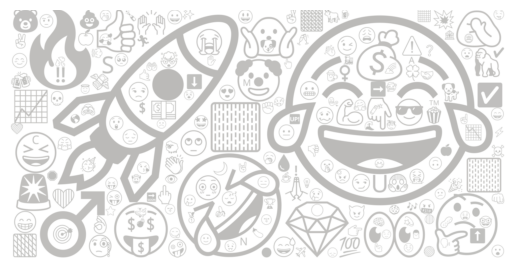}
    }
    \caption{Emoji Clouds for Financial Microblogs. (a) Bullish (green\rev{/left}), (b) Bearish (red\rev{/middle}), and (c) Unlabeled (gray\rev{/right}) emoji clouds, with emoji size reflecting frequency of use.}
    \label{fig:clouds}
\end{figure}

Fig.~\ref{fig:clouds} highlights differences in emoji usage between bullish and bearish posts. For instance, the \emoji{rocket} emoji appears significantly more frequently (larger) in the bullish emoji cloud, whereas the \emoji{pile-of-poo} emoji is more prominent in the bearish cloud. Additionally, the laughing emojis \emoji{joy} and \emoji{rofl} are prevalent in both clouds but are more pronounced in bearish posts. This could suggest that sarcasm, as indicated by the use of these emojis, is more common in bearish sentiments.

The gray cloud represents posts that have not been labeled by the poster, possibly indicating uncertainty. Emojis commonly associated with uncertainty, such as \emoji{shrug} and \emoji{thinking}, are more prominent in the unlabeled emoji cloud compared to the bullish or bearish clouds. However, the presence of unlabeled posts does not necessarily imply uncertainty about sentiment. The frequent use of sentiment-specific emojis like \emoji{rocket} in the gray cloud indicates that some users may be confident in their sentiment but opt not to label their posts explicitly.

As shown in Table~\ref{table:tokenlengthsPercentile}, the number of words in text-only data is significantly higher than the number of emojis in emoji-only data. Therefore, the information contained in emojis of a post can be analyzed by a model with significantly fewer tokens. This allows the analysis and inference to be faster and more efficient. The sentiment of a post containing many textual tokens might be summarized in just one emoji token. In sentiment analysis, emojis offer a lower lexical diversity and a smaller feature space compared to text due to their limited set of symbols. There are many fewer unique emojis than there are unique words. In our text-only data, there are around 227,000 unique words. In contrast, the emoji-only data contains fewer than 3,000 unique emojis. This smaller dimensionality simplifies machine learning models, resulting in less computational and data demands.

\begin{table}[ht]
\centering
\caption{\label{table:tokenlengthsPercentile}Percentile Distribution of Text and Emoji Lengths in Posts. This table displays the character length distribution for text-only and emoji-only content across various percentiles, demonstrating the comparative brevity of emoji usage.}
\begin{tabular}{lccccccc}
\toprule
\textbf{Percentile} & 5$^\mathrm{th}$ & 25$^\mathrm{th}$ & 50$^\mathrm{th}$ & 75$^\mathrm{th}$ & 90$^\mathrm{th}$ & 95$^\mathrm{th}$ & 99$^\mathrm{th}$ \\ \midrule
\textbf{Text Length} & \rev{1} & \rev{20} & \rev{45} & \rev{90} & \rev{165} & \rev{248} & \rev{484} \\ 
\textbf{Emoji Length} & \rev{1} & \rev{1} & \rev{2} & \rev{4} & \rev{7} & \rev{11} & \rev{24} \\ \bottomrule
\end{tabular}
\end{table}

\section{Emoji Usage: StockTwits versus $\mathbb{X}$ (Twitter)}
\label{sec:emoji_usage_StockTwits_versus_twitter}

In this section, we compare emoji usage on two platforms: StockTwits and $\mathbb{X}$ (Twitter).

\begin{figure}[ht]
    \centering
    \subfloat[Twitter ($\mathbb{X}$)]{
        \includegraphics[width=0.48\textwidth]{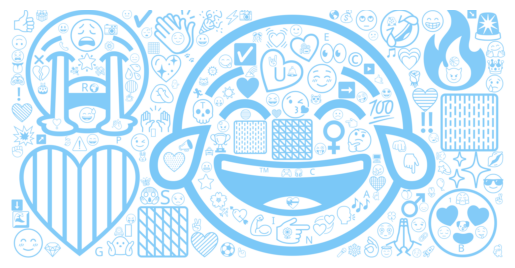}
    }
    \subfloat[StockTwits]{
        \includegraphics[width=0.48\textwidth]{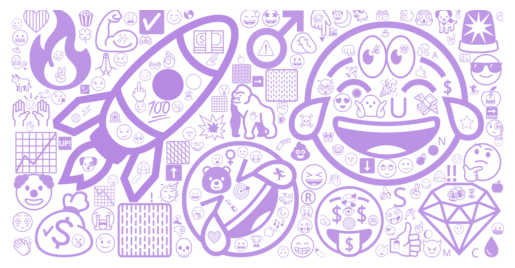}
    }
    \caption{Emoji Clouds for Twitter ($\mathbb{X}$) and StockTwits. (a) Twitter ($\mathbb{X}$) (blue\rev{/left}) and (b) StockTwits (purple\rev{/right}) emoji clouds visualize the frequency of emoji use on each platform. The size of each emoji indicates its relative frequency, showcasing distinct preferences for emojis on both platforms.}
    \label{fig:Twitterclouds}
\end{figure}

Fig.~\ref{fig:Twitterclouds} shows emoji clouds of Twitter (blue\rev{/left}, now $\mathbb{X}$) and of StockTwits (purple\rev{/right}). Looking at the clouds, noticeable differences in emoji usage between the two platforms are immediately apparent. For instance, Twitter's cloud contains more heart-related emojis (such as \emoji{heart} or \emoji{heart-eyes}), while StockTwits' contains more money-related emojis (such as \emoji{moneybag} or \emoji{money-mouth-face}). Different emojis are also much more prevalent on one platform than the other. For example, \emoji{rocket} and \emoji{gem-stone} are much bigger in the StockTwits' cloud compared to Twitter's. However, some emojis seem to be used equally frequently on both platforms, particularly \emoji{joy} and \emoji{fire}.

From Table \ref{StockTwitsRanking}, it is evident that the distribution of the 20 most used emojis on StockTwits likely differs from that on Twitter, particularly due to the prevalence of financially themed emojis, such as \emoji{money-mouth-face}. Supporting this, both the Mann-Whitney U and the Kolmogorov-Smirnov tests indicate a highly significant difference in distributions ($p$-value < 0.01). \rev{Supporting this, a chi-square test confirms a highly significant difference between platforms ($p<0.01$), with a large effect size (Cram\'{e}r’s $V=0.599$).} This finding underscores the need for context-specific sentiment analysis models tailored to financial social media, aligning with extensive research in the field that advocates for such approaches~\cite{du2024financial, renault2020sentiment, nasekin2020deep, wilksch2023pyfin, kulakowski2023sentiment}.

\begin{table}[ht]
\centering
\caption{\label{StockTwitsRanking}Comparative Analysis of Emoji Usage on StockTwits and Twitter ($\mathbb{X}$). This table ranks the top 20 emojis used on both platforms, highlighting differences in frequency and specific emoji preferences.}
\begin{adjustbox}{width=1\textwidth,center}
\Huge
\begin{tabular}{|*{22}{c|}}
\hline
 & & \emoji{rocket} & \emoji{joy} & \emoji{rofl} & \emoji{fire} & \emoji{gem} & \emoji{money-mouth-face} & \emoji{moneybag} &\emoji{eyes} & \emoji{thinking} & \emoji{smiling-face-with-sunglasses} & \emoji{chart-with-upwards-trend} & \emoji{thumbsup} &\emoji{bear} & \emoji{clown-face} &\emoji{gorilla} &\emoji{muscle} &\emoji{raised-hands} &\emoji{grin} &\emoji{laughing} &\emoji{boom}
 \\ \hline
\multirow{2}{*}{Rank} & StockTwits & 1 & 2 & 3 & 4 & 5 & 6& 7& 8& 9& 10& 11& 12& 13& 14& 15& 16& 17& 18& 19& 20\\ \cline{2-22}
 & Twitter & 97 & 1 & 13 & 5 & 71 & 215 & 80 & 17 & 15 & 24 & 281 & 31 & 233 & 358 & 664.5 & 47 & 34 & 26 & 74 & 35\\ \hline
\multirow{2}{*}{Rel. Freq.} & StockTwits & 15.78\% & 12.3\% & 6.82\% & 4.95\% & 4.80\% & 3.37\% & 3.28\% & 2.87\%& 2.67\%& 2.59\%& 2.32\%& 2.27\% & 2.19\%& 1.76\%& 1.64\%& 1.63\%& 1.52\%& 1.47\% & 1.43\% & 1.34\% \\ \cline{2-22}
& Twitter & 0.31\% & 13.27\% & 1.41\% & 4.14\%& 0.44\% & 0.12\% & 0.37\%& 1.29\%& 1.30\%& 0.90\%& 0.09\%& 0.83\%& 0.11\%& 0.06\%& 0.02\%& 0.59\%& 0.75\%& 0.86\%& 0.39\% & 0.73\%\\ \hline
\end{tabular}
\end{adjustbox}
\end{table}



\section{Emoji Sentiment Scores}
\label{sec: emoji_sentiment_scores}
In this section, we score emojis based on their presence in bullish versus bearish posts. For every emoji, we calculate the percentage of posts that are bullish versus bearish when that particular emoji is present.

\begin{figure}[ht]
   \centering
   \includegraphics[width=\linewidth]{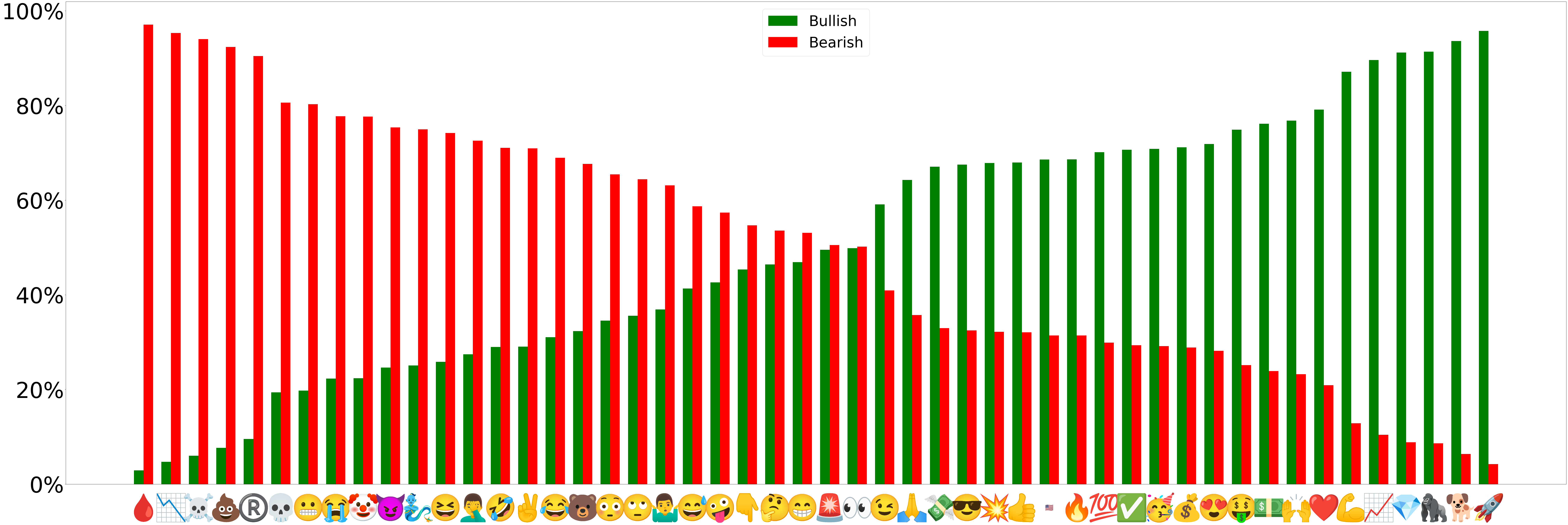}
   \caption{Sentiment Score of Individual Emojis. This bar chart displays the proportion of bullish (green) and bearish (red) posts associated with the 50 most used emojis, highlighting specific emojis as strong predictors of market sentiment.}
    \label{fig:emojiscores}
\end{figure}

The bars in Fig.~\ref{fig:emojiscores} represent the percentage of posts that are bullish/bearish, given that an emoji is present. As shown in Fig.~\ref{fig:emojiscores}, some emojis can serve as strong predictors of the sentiment for a post. For instance, in our sample, 96\% of posts that contain \emoji{rocket} are bullish, and 97\% of posts that contain \emoji{drop-of-blood} are bearish. Such findings highlight the predictive value some emojis have in financial sentiment analysis, enabling highly efficient sentiment classification.

\begin{figure}[ht]
   \centering
   \includegraphics[width=\linewidth]{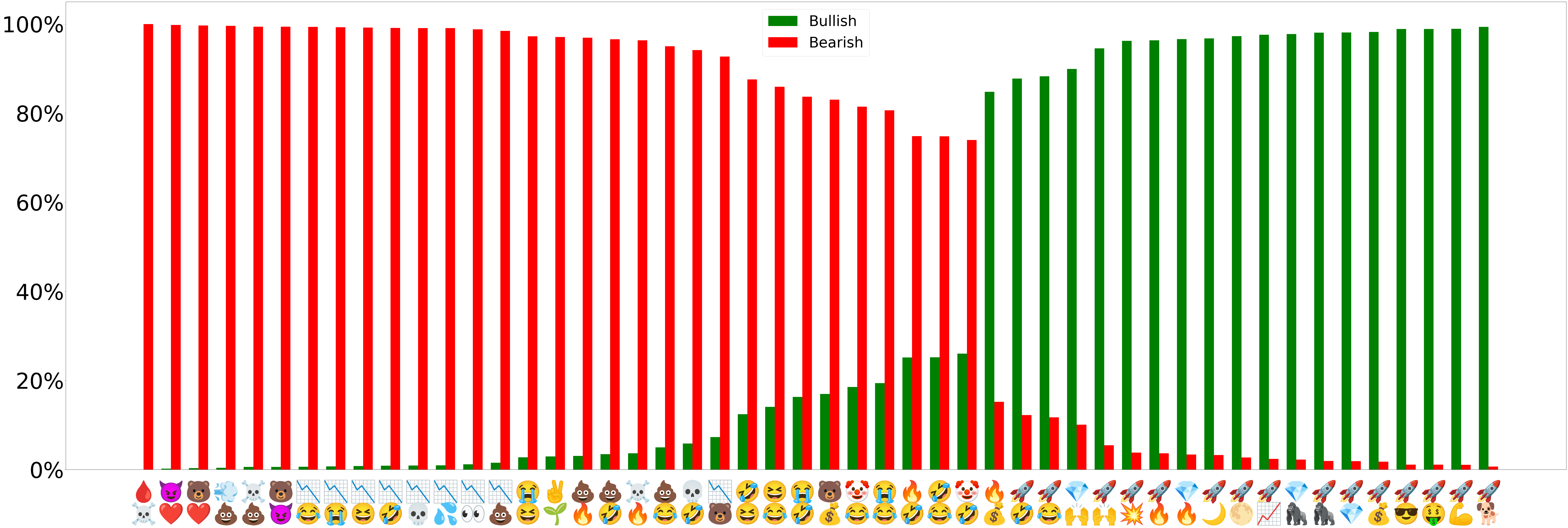}
   \caption{Sentiment Score of Emoji Pairs. This bar chart displays the proportion of bullish (green) and bearish (red) posts associated with the 50 most used emoji pairs.}
   \label{fig:emojiscoresPair}
\end{figure}

Fig.~\ref{fig:emojiscoresPair} reveals that most emojis retain their sentiment scores whether analyzed individually or as part of a pair. However, there are notable exceptions. The \emoji{heart} emoji is typically indicative of positive sentiment when analyzed alone, but also appears in two of the most bearish pairs. A similar observation applies to the \emoji{fire} and \emoji{money-bag} emojis to some extent, given the bearish combinations of \emoji{poop} with \emoji{fire} and \emoji{bear} with \emoji{money-bag}. In the first case, the \emoji{heart} may just be used in the meaning of ``to like'', potentially disclosing a short position of the original poster. In the second case, \emoji{poop} and \emoji{bear} act as negations, figurative speech, or qualitative modifiers, e.g., ``\emoji{bear} on \emoji{fire}''. It is worth pointing out in this context that such modifiers (especially negations) in natural language historically posed challenges for NLP. The reason is that they necessitated either longer attention (in the case of transformers) or longer N-grams (in the case of Bag-of-Words approaches).

The finding that an individual or pair of emojis is highly bullish or bearish can be used to create efficient sentiment analysis algorithms. We can have a level of confidence that if an emoji or pair of emojis exists in a post, then it is either bullish/bearish. The sharp transition between bullish and bearish pairs also signals a low amount of ambiguity regarding the sentiment of emoji pairs.

Additionally, some emojis, such as the \emoji{thinking-face}, are frequently used alone and so do not appear in the most common pairs. Moreover, the prevalence of laughing emojis like \emoji{joy} and \emoji{rofl} in bearish pairs indicates they may be used to convey sarcasm, mock optimism, or express self-deprecating humor. We also notice that the score for pairs is skewed towards bearish. This raises the question of whether the number of emojis in a post can also be predictive of its sentiment.

\begin{figure}[ht]
   \centering
   \includegraphics[width=\linewidth]{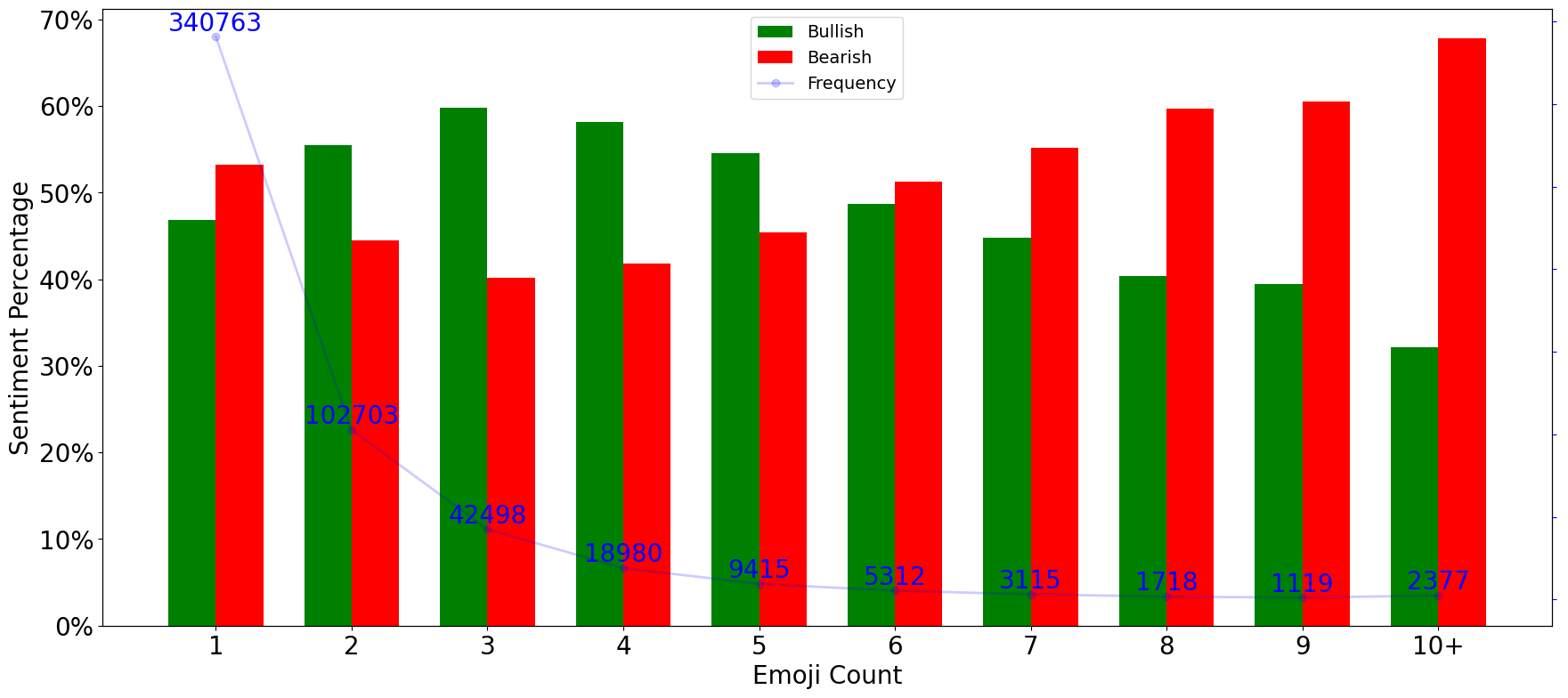}
   \caption{Emoji Count versus Sentiment and Frequency. This chart displays the percentage of bullish (green) and bearish (red) posts based on the number of unique emojis used, along with the frequency (blue line) of posts for each emoji count.}
    \label{fig:numEmojis}
\end{figure}

Fig.~\ref{fig:numEmojis} displays the distribution of bullish and bearish sentiment across posts categorized by the number of emojis used. Posts containing two to five emojis exhibit a higher likelihood of expressing bullish sentiment, with a peak observed at three emojis. However, as the number of emojis surpasses five, a clear shift towards bearish sentiment emerges, culminating in posts with ten or more emojis demonstrating the highest proportion of bearish sentiment (this could be considered a ``pictogram rant''). The frequency of posts decreases with increasing emoji counts, resembling an exponential decay pattern. 

All these insights gained from analyzing the use and distribution of emojis in these posts lead us to hypothesize that emojis may carry a meaning strong enough to classify sentiment using relatively shallow or simple models. Consequently, we opted to also include logistic regression in our experiments, aiming for high inference speed that is beneficial for sentiment analysis in high-frequency trading, where even microseconds matter~\cite{aquilina2022quantifying}.

\section{Emoji Sentiment Analysis Algorithms}
\label{sec:sentiment_analysis_algorithms}
In this section, we perform sentiment analysis using only emojis.

\subsection{Logistic Regression}
\label{subsec:logistic}

Table~\ref{logisticPerformance} presents the accuracy and speed metrics for logistic regression, comparing using text only, emoji only, and text plus emojis. As expected, text plus emojis performs the best, with an F$_1$ score of 0.82. It is worth noting, though, that including text (as opposed to only emojis) in the sentiment analysis is only able to boost the F$_1$ score by 7 percentage points (up from 0.75). In addition, using only text fares marginally better (0.76) than using only emojis. This supports our hypothesis that emojis in financial social media carry strong meaning in an extremely compact form. Furthermore, this accuracy is surprisingly good, given the simplicity and speed of the logistic regression model. 

The higher precision metric of the emojis-only model (compared to the text-only) means that it is more likely to be accurate when labeling a post as bullish, but the lower recall metric means that it is more likely to miss posts that are bullish. The trade-off between precision and recall can be controlled by changing the decision threshold. \rev{However, for the sake of easier reproducibility, we kept the default decision threshold of 0.5 for all models.}

As expected, using only emojis has an advantage in terms of computational efficiency. Another major benefit of using emojis in financial social media is that they are arguably more generalizable across cultures compared to text and, thus, could be valuable for cross-language generalization (left for future work).

\begin{table}[ht]
\centering
\caption{\rev{\label{logisticPerformance}Performance of Logistic Regression Models}. This table compares the accuracy and speed of logistic regression models trained on text only, emojis only, and text with emojis. Time is measured in seconds. \rev{We report average and 95\% confidence intervals in square brackets.} 
Both training and inference processes were executed on a 13$^\mathrm{th}$ Gen Intel Core i9-13900H laptop CPU with 32GB of RAM.}
\begin{tabular}{lccc}
\toprule
 & \textbf{Text only} & \textbf{Emojis only} & \textbf{Text + Emojis} \\ 
\midrule
\textbf{Accuracy Metrics} & & & \\
Recall & 0.762 [0.758, 0.766] & 0.754 [0.750, 0.757] & 0.824 [0.821, 0.827] \\
Precision & 0.762 [0.758, 0.765] & 0.755 [0.752, 0.759] & 0.824 [0.820, 0.827] \\
F$_1$ Score & 0.761 [0.757, 0.765] & 0.754 [0.750, 0.758] & 0.824 [0.821, 0.827] \\ 
\midrule
\textbf{Speed Metrics} & & & \\
Training time (s) & 4.106 [3.870, 4.310] & 3.400 [3.218, 3.586] & 4.654 [4.486, 4.961] \\
Inference time (s) & 0.003 [0.002, 0.003] & 0.001 [0.001, 0.002] & 0.003 [0.002, 0.003] \\ 
\bottomrule
\end{tabular}
\end{table}

\begin{figure}[ht]
    \centering
    \begin{subfigure}[b]{0.3\textwidth}
        \includegraphics[width=\textwidth]{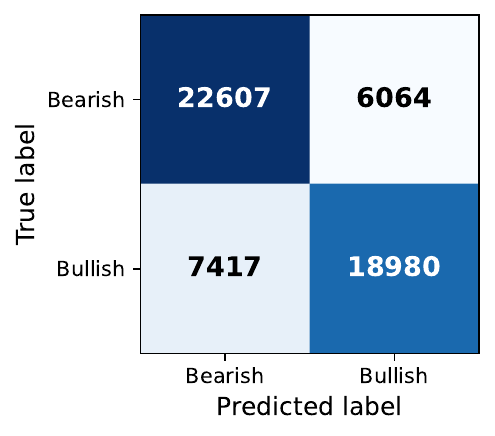}
        \caption{Emoji Only}
    \end{subfigure}
    \hfill 
    \begin{subfigure}[b]{0.3\textwidth}
        \includegraphics[width=\textwidth]{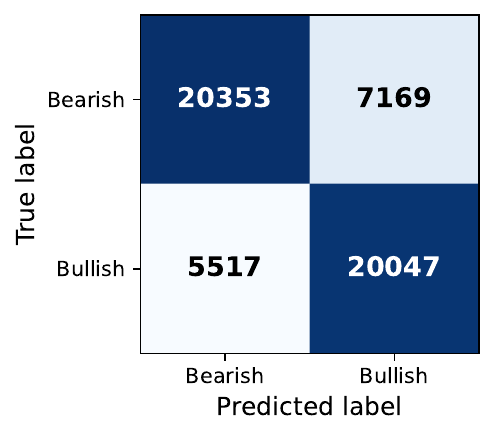}
        \caption{Text Only}
    \end{subfigure}
    \hfill
    \begin{subfigure}[b]{0.3\textwidth}
        \includegraphics[width=\textwidth]{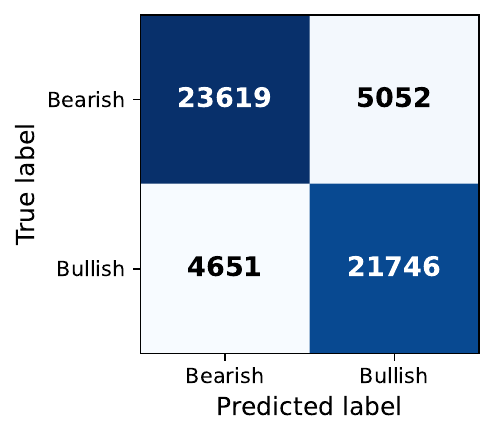}
        \caption{Text and Emoji}
    \end{subfigure}
    \caption{Confusion Matrices of Logistic Regression Models. This figure displays confusion matrices for logistic regression models trained on emojis only (a), text only (b), and text with emojis (c). The matrices compare the models' predicted sentiment labels (bullish or bearish) with the true labels.}
\end{figure}

\begin{figure}[ht]
   \centering
   \includegraphics[width=\linewidth]{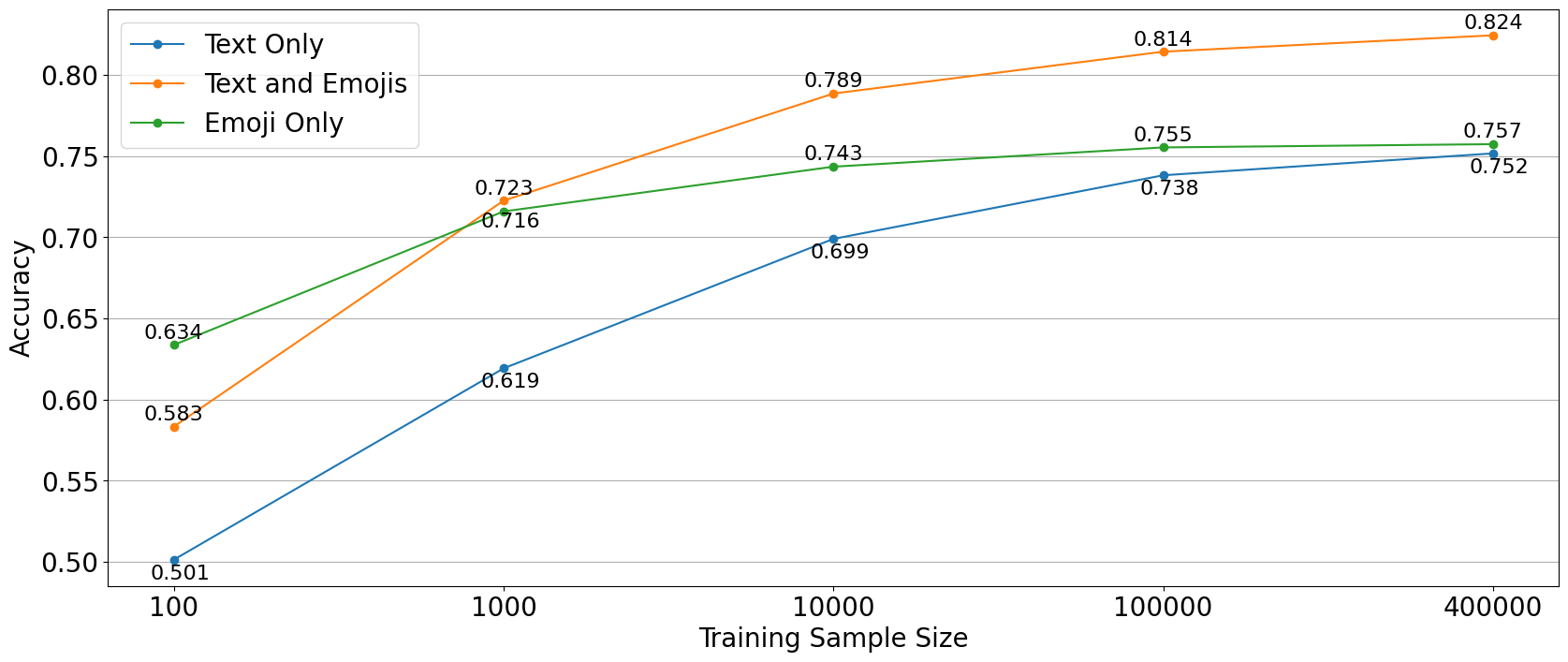}
   \caption{Impact of Training Sample Size on Logistic Regression Model Accuracy. This figure illustrates the relationship between training sample size and model accuracy for logistic regression models trained on text only (blue line), emojis only (green line), and text with emojis (orange line). Accuracy is evaluated on the same separate test set of 50,000 posts.}
   \label{fig:sampleSize}
\end{figure}

Fig.~\ref{fig:sampleSize} depicts the model accuracy as a function of training data size. With its logarithmic X-axis scale, the figure clearly demonstrates ``diminishing returns''---the increase in accuracy decreases as the training data size increases significantly, approaching a horizontal asymptote of the theoretical maximum accuracy possible with the given models.

When the model is trained on text only, the accuracy starts at 50.1\% for 100 posts and increases as more data is used, reaching approximately 75.2\% accuracy with 400,000 posts.
The model trained on only emojis starts with 63.4\% accuracy for 100 posts and also demonstrates a steeper curve than text only, achieving about 75.7\% accuracy with 400,000 posts.
The model trained with both text and emojis starts with a higher accuracy of 58.3\% at 100 posts and shows a steeper increase in accuracy as the sample size grows, reaching about 82.4\% with 400,000 posts.

One of the advantages of using emojis only, as shown in Fig.~\ref{fig:sampleSize}, is the ability to train on a lower amount of data. In this figure, we see that the logistic regression model, when trained on only 1,000 posts, achieves an accuracy of around 72\% on a test sample of 50,000 posts, indicating a potential advantage in contexts where training data is limited. In contrast, a text-only model requires over 10,000 posts to surpass 70\% accuracy. This discrepancy highlights the efficiency of using emojis for sentiment analysis, as they allow for achieving high accuracy with fewer training samples.

\begin{table}[h]
\centering
\caption{\rev{Entropy and Vocabulary Statistics (Top 90\% Frequency Mass)}}
\label{tab:entropy_results}
\begin{tabular}{l r}
\hline
\textbf{Metric} & \textbf{Value} \\
\hline
Word entropy  & 9.52 bits \\
Emoji entropy  & 6.72 bits \\
Average word length & 5.48 characters \\
Unique words & 5,042 \\
Unique emojis & 267 \\
\hline
\end{tabular}
\end{table}

\rev{Because emojis’ efficiency advantages relate to how compactly information is encoded, entropy measurements provide a natural way to compare text and emojis. As shown in Table~\ref{tab:entropy_results}, the top-90\% word distribution yields an entropy of 9.52 bits. This value is lower than Shannon’s classical estimate of 11.82 bits per word for printed English \cite{shannon1951prediction}. A lower entropy reflects a more homogeneous language in the shorter-form financial social media compared to printed English. The average word length in our dataset is 5.48 characters, slightly higher than Shannon’s reference value of 4.5 letters but in line with other studies on social media text \cite{smirnov2017digital}}. 

\rev{In contrast, emoji usage is more concentrated, with the top-90\% mass consisting of 267 emojis and producing an entropy of 6.72 bits. The corresponding top-90\% vocabulary for words is much larger (5,042 words), reflecting the broader range of textual content relative to emojis.}



\subsection{Transformer}
\label{subsec:Transformer}

Table~\ref{transformerPerformance} presents the performance of a transformer~\cite{vaswani2017attention} model on text only, emojis only, and both text and emojis datasets. Similar to the logistic regression model shown in Table~\ref{logisticPerformance}, the transformer model demonstrates the highest accuracy when analyzing data that includes both text and emojis. However, unlike the logistic regression model, the transformer shows better performance with text-only data than with emoji-only data. This indicates that the transformer model was able to capture more information about the meaning of text, which is presumably more complex than capturing information about the meaning of emojis. This is likely because short emoji sequences do not fully exploit the transformer’s attention mechanism.

The relative simplicity of emoji analysis is further highlighted by the comparable performance of the simpler logistic regression model and the more complex transformer model on emoji-only data. For emojis-only, using transformers may be excessive, as simpler models can achieve similar levels of accuracy with greater efficiency.



\begin{table}[ht]
\centering
\caption{\label{transformerPerformance}\rev{Performance of the Twitter-RoBERTa Model}. This table compares the accuracy and speed of the model when trained on text only, emojis only, and text with emojis. Time is measured in seconds. \rev{We report average and 95\% confidence intervals in square brackets.} 
Both training and inference processes were executed on a 13$^\mathrm{th}$ Gen Intel Core i9-13900H laptop CPU with 32GB of RAM.}
\begin{tabular}{lccc}
\toprule
 & \textbf{Text only} & \textbf{Emojis only} & \textbf{Text + Emojis} \\ 
\midrule
\textbf{Accuracy Metrics} & & & \\
Recall & 0.841 [0.838, 0.844] & 0.763 [0.759, 0.766] & 0.884 [0.880, 0.886] \\
Precision & 0.841 [0.838, 0.844] & 0.765 [0.761, 0.768] & 0.884 [0.881, 0.887] \\
F$_1$ Score & 0.841 [0.838, 0.844] & 0.762 [0.758, 0.766] & 0.883 [0.880, 0.886] \\ 
\midrule
\textbf{Speed Metrics} & & & \\
Training time (s) & 93 [92, 97] & 95 [94, 95] & 95 [95, 97] \\
Inference time (s) & 19 [18, 28] & 22 [21, 24] & 22 [21, 26] \\ 
\bottomrule
\end{tabular}
\end{table}

\begin{figure}[ht]
    \centering
    \begin{subfigure}[b]{0.3\textwidth}
        \includegraphics[width=\textwidth]{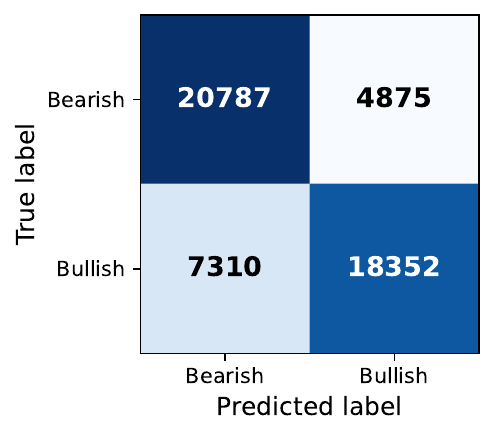}
        \caption{Emoji Only}
    \end{subfigure}
    \hfill 
    \begin{subfigure}[b]{0.3\textwidth}
        \includegraphics[width=\textwidth]{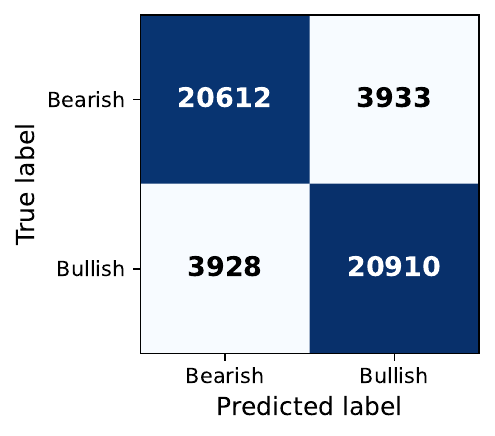}
        \caption{Text Only}
    \end{subfigure}
    \hfill
    \begin{subfigure}[b]{0.3\textwidth}
        \includegraphics[width=\textwidth]{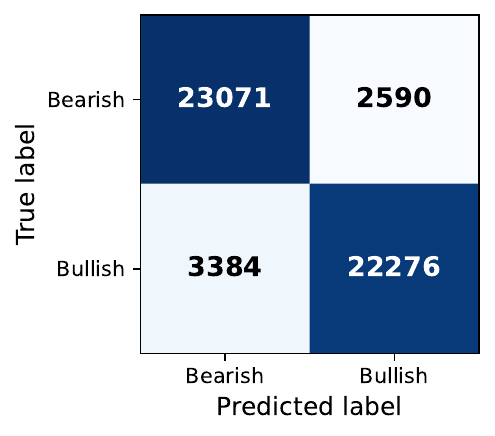}
        \caption{Text and Emoji}
    \end{subfigure}
    \caption{Confusion Matrices of Transformer-based Twitter-roBERTa Models. This figure displays confusion matrices for Twitter-roBERTa models trained on emojis only (a), text only (b), and text with emojis (c). The matrices compare the models' predicted sentiment labels (bullish or bearish) with the true labels.}
\end{figure}

Similar to the results shown for the regression model in Fig.~\ref{fig:sampleSize}, the transformer model demonstrates higher accuracy on emoji-only data, even when trained with a relatively small dataset (1,000 posts) and tested on 50,000 posts. However, in contrast to the regression model, the transformer's complexity allows it to derive greater benefits from a larger text dataset. This advantage is reflected in the increasing accuracy observed in the text-only model (blue line).

One of the hyperparameters that can have a significant impact on training/inference speed is the context length (that is, the maximum number of tokens used to predict the next token). 
Naturally, having longer contexts requires more computational resources and memory. As shown in Table~\ref{tab:max-length}, we see that, quite intuitively, emoji-only data requires smaller context lengths to achieve its highest accuracy. For instance, emoji-only data reaches its highest accuracy in a context of 10 or 20 tokens, while text-only data requires 50 tokens. This is expected as, in general, posts contain significantly fewer emojis than text (cf. Table~\ref{table:tokenlengthsPercentile}). However, the speed decreased less than expected. For 3 to 20 tokens, the speed is mostly constant, decreasing slightly for 50 tokens and almost halving for 120 tokens. The self-attention mechanism at the core of transformer models exhibits quadratic time and space complexity relative to input length~\cite{keles2023computational}. Therefore, increasing the maximum token length from 3 to 120 should increase the time complexity by a factor of 1,600. However, the increase in practice can be significantly different. The reason is that modern computing architectures tend to be severely limited by memory bandwidth. Various levels of caches are used to improve memory bandwidth at a reasonable cost to hardware vendors, but cache sizes remain generally limited. As a result of a transformer requiring additional memory depending linearly on the context length (``KV cache''), smaller context lengths benefit disproportionately from such hardware caches.
As shown in Table~\ref{tab:max-length}, increasing the context length from 3 to 120 increases the time needed to train and infer by only approximately $2{\times}$. 


\begin{table}[ht]
\centering
\caption{\label{tab:max-length}Effect of Context Length on Model Performance and Speed. This table demonstrates the impact of varying sequence max-length on the F$_1$ score, training speed, and inference speed of Twitter-RoBERTa models trained on text-only and emoji-only data. Time is measured in seconds. Both training and inference processes were executed on a 13$^\mathrm{th}$ Gen Intel Core i9-13900H laptop CPU with 32GB of RAM.}
\begin{tabular}{lcccccc}
\toprule
                  \textbf{Sequence Max-Length}  & \textbf{3} & \textbf{10} & \textbf{20} & \textbf{50} & \textbf{120} \\ \midrule
\textbf{F1 Score}   &            &             &             &             &              \\
Text-Only          & 0.64       & 0.76        & 0.81        & 0.83        & 0.82         \\
Emoji-Only         & 0.64       & 0.74        & 0.76        & 0.75        & 0.76         \\ \midrule
\textbf{Training Speed} &        &             &             &             &              \\
Text-Only          & 272        & 266         & 275         & 291         & 510          \\
Emoji-Only         & 264        & 266         & 271         & 286         & 469          \\ \midrule
\textbf{Inference Speed} &       &             &             &             &              \\
Text-Only          & 18         & 18          & 19          & 20          & 37           \\
Emoji-Only         & 18         & 18          & 17          & 19          & 34           \\ \bottomrule
\end{tabular}
\end{table}

\subsection{Other Tested Models}
\label{subsec:other_models}

In this section, we explore the performance of various machine learning models on the emoji dataset as potential baseline models against the more complex transformer model. These models were selected for their relative simplicity and computational efficiency, which could offer practical advantages in financial sentiment analysis. \rev{Here, simplicity refers to the model architecture and number of trainable parameters compared to the heavier deep-learning transformer models tested, while computational efficiency is evaluated empirically using training time and inference time as reported in Table~\ref{table:other_models}.}

Table~\ref{table:other_models} indicates that all models achieve an accuracy (F$_1$ scores) competitive with the transformer model when using emoji-only data, while offering much faster speeds and greater computational efficiency. Among the baseline models---we used Multinomial Na\"{i}ve Bayes (MNB), Random Forest (RF), Light Gradient Boosting Machine (LGBM), Linear Support Vector Classification (SVC), and Logistic Regression (Reg),---we generally observe uniform accuracy (F$_1$ Score), indicating robust effectiveness across different modeling techniques. 

However, there are some significant differences in training and inference times among the tested models. RF requires significantly more time to train than the other models, making it the least efficient in terms of speed. LGBM also shows longer training times, but it is faster than RF. MNB, SVC, and Reg are all relatively fast, with MNB being the fastest, approximately ten times faster than training Reg.

\begin{table}[ht]
\centering
\caption{\label{table:other_models}Performance Comparison of Baseline Machine Learning Models on Emoji Data. This table evaluates the accuracy and speed of various baseline models, including Multinomial Na\"{i}ve Bayes (MNB), Random Forest (RF), LightGBM (LGBM), and LinearSVC (SVC), in comparison to the Logistic Regression (Reg) and Transformer (Tr) models, all using emoji-only data. Time is measured in seconds. Both training and inference processes were executed on a 13$^\mathrm{th}$ Gen Intel Core i9-13900H laptop CPU with 32GB of RAM.}
\begin{tabular}{lcccccc}
\toprule
                    & \textbf{MNB} & \textbf{RF} & \textbf{LGBM} & \textbf{SVC} & \textbf{Reg} & \textbf{Tr}\\ \midrule
\textbf{Accuracy} &&&&&&\\
Recall &0.72&0.72&0.68&0.72&0.72&0.73\\
Precision &0.77&0.79&0.80&0.77&0.78&0.79\\
F1 Score &0.74&0.76&0.74&0.75&0.75&0.76\\ \midrule
\textbf{Speed} &                   &        &              &&&                        \\
Training time &0.03&31.91&2.76&0.57&0.29&498\\
Inference time &0.003&0.58&0.04&0.002&0.001&38\\ \bottomrule
\end{tabular}
\end{table}

\subsection{Huggingface Model Comparisons}
\label{subsec:huggingfaceModelsComparison}

In this section, we evaluate various transformer models from Huggingface, focusing on their application in financial social media sentiment analysis. This analysis was performed in order to choose the best transformer model to test in Subsection~\ref{subsec:Transformer}.

Each model reviewed in Table~\ref{table:huggingfacemodels} is based on the BERT~\cite{BERT} architecture, a well-known implementation of the transformer framework.

Twitter-RoBERTa \cite{loureiro2022timelms} is a model designed for sentiment analysis of general Twitter posts. It is pre-trained on approximately 124 million tweets and fine-tuned on the TweetEval dataset (which includes more than 100,000 manually annotated tweets~\cite{rosenthal2019semeval}). 

FinBERT \cite{araci2019finbert}
is a model designed for sentiment analysis of financial news articles. It pre-trains the BERT model
on 1.8 million financial news articles published by Reuters (TRC2\footnote{\url{https://trec.nist.gov/data/reuters/reuters.html}}). It is then fine-tuned on the Financial PhraseBank dataset\footnote{\url{https://github.com/vrunm/Text-Classification-Financial-Phrase-Bank}}, which consists of 4,845 financial news articles with manually annotated sentiment. By the nature of the data, this model has probably not seen any emojis before. 

DistilBERT \cite{sanh2019distilbert} is a general language model designed to output the same probabilities as the BERT model, but it is smaller and faster. It is pre-trained on BookCorpus\footnote{\url{https://yknzhu.wixsite.com/mbweb}} and the English Wikipedia\footnote{\url{https://huggingface.co/datasets/wikipedia}}. Using this model for sentiment analysis requires the addition of a classification layer, as it is not designed for text classification. This is accomplished with the \texttt{DistilBertForSequenceClassification} class from the Hugging Face transformers library. By the nature of the data, this model has probably not seen any emojis before. 

FinancialBERT \cite{hazourli2022financialbert} is a model designed for sentiment analysis of formal financial texts. It pre-trains the BERT model on financial news articles published by Reuters and Bloomberg, combined with a variety of corporate reports and earnings call transcripts. Similar to FinBERT, it then fine-tunes the pre-trained model on the Financial PhraseBank dataset. By the nature of the data, this model has probably not seen any emojis before. 


CryptoBERT\footnote{\url{https://huggingface.co/kk08/CryptoBERT}} is a model designed for sentiment analysis of cryptocurrency-related texts. This is achieved by fine-tuning FinBERT using a small corpus of labeled cryptocurrency-related texts. 

FinTwitBERT \cite{FinTwitBERT} is a model designed for the sentiment analysis of financial tweets. It is pre-trained on 10 million financial tweets and then fine-tuned on around 40,000 human-labeled financial tweets (which are augmented by over 1 million synthetic tweets). It seems that this model has not incorporated emojis properly during fine-tuning, as evidenced by its performance on emojis; for instance, it has failed to classify a post consisting of only several rockets (\emoji{rocket}) as bullish, as a good model would. 



\begin{table}[ht]
\centering
\caption{F$_1$ Scores of Fine-tuned Transformer-based BERT Models. This table presents the F$_1$ scores achieved by various BERT models when trained on text only, emojis only, and text with emojis.}
\label{table:huggingfacemodels}

\begin{tabular}{lccc}
\toprule
                    \textbf{F1 Score} & \textbf{Text only} & \textbf{Emojis only} & \textbf{Text + Emojis} \\ \midrule
\textbf{Model} &                   &                      &                        \\
Twitter-RoBERTa & 0.84                  &0.76                      &0.88                        \\
FinBERT & 0.82                  &0.69                      &0.87                        \\
DistilBERT  & 0.81                  &0.69                      &0.87                        \\
FinancialBERT &0.80                   &0.69                      &0.86                        \\
CryptoBERT &0.81                   &0.69                      &0.87                        \\
FinTwitBERT &0.81                   &0.69                      &0.87                        \\\bottomrule
\end{tabular}
\end{table}

It is noticeable that the models perform generally similarly, except for Twitter-RoBERTa. The similarity in performance signals the existence of a sharp decision boundary that all models converge to after fine-tuning. The models are also all derived from the same BERT architecture. Without fine-tuning them on our data, the models achieve significantly more varying accuracy; this variation is possibly due to the data they have been trained on initially, which does not include any social media or financial data for some of the models. Note that we fine-tuned and tested all models using the exact same data sample.

Some of the models remove emojis during the tokenization process. For them, we demojized the emojis. Demojizing emojis means converting emojis to their textual aliases or descriptions \footnote{\url{https://pypi.org/project/emoji/}}. 
The best-performing model was Twitter-RoBERTa, which is the only model tested that does not remove emojis during tokenization. The performance of Twitter-RoBERTa with and without demojizing is similar. So, it seems that the increased performance is specific to the way the model was pre-trained, taking emojis into consideration, while the other models excluded emojis.

For example, when processing a sentence containing multiple emojis, the tokenizer for Twitter-RoBERTa retains and correctly interprets emojis. After encoding these into numbers and then decoding them back to the text, the sentence is rendered as: \texttt{<s>\emoji{mag}\emoji{eyes} different tokenizers \emoji{input-symbols}\emoji{arrow-right}\emoji{input-numbers} emojis differently.</s>}. In contrast, the tokenizers from all the other models tested failed to recognize emojis. After their encoding and subsequent decoding, they replace emojis with a placeholder token \texttt{[UNK]} (``unknown''), resulting in the following processed output for the above input: \texttt{[CLS] [UNK] different tokenizers [UNK] emojis differently. [SEP]}. 
This indicates that these models are either not trained to handle emojis or are designed to ignore them, possibly because they focus on data that does not include emojis, such as formal financial text or English books.










\section{Discussion}
\label{sec: discussion}

This section discusses the main results of the paper, highlights implications, acknowledges limitations, and outlines future research directions.

\subsection{Results}
In this study, we examined the role of emojis in financial sentiment analysis. Our objective was to determine the effectiveness and efficiency of using emojis for sentiment analysis in financial contexts, assessing aspects like accuracy performance, computational efficiency, data requirements, and usage patterns. Our research used data from 18.5 million posts, involving both a descriptive analysis of emoji usage and a comparative analysis using different sentiment analysis models. Our exploration revealed several key insights. While integrating both text and emojis yielded the highest accuracy in sentiment classification, using emojis alone demonstrated effective accuracy, with the added benefit of significantly lower data requirements and computational costs. Specific emojis and emoji pairs exhibited strong predictive power (above 90\%) for bullish or bearish sentiment, highlighting their potential for efficient and rapid sentiment analysis. Our analysis also quantified the extreme difference in emoji usage between financial and general social media, emphasizing the need for domain-specific sentiment analysis models. Another surprising finding is that models trained on just 1,000 posts can achieve relatively high accuracy when emojis are considered, even when tested on a larger sample of 50,000 posts, highlighting the training dataset-size advantage working with emojis offers for sentiment analysis.

By utilizing transformer models and properly incorporating emojis, we achieved a sentiment classification accuracy of 0.88 F$_1$ score, surpassing the performance reported in several previous studies. Furthermore, our experimentation with various models, including traditional machine learning algorithms and transformer-based models, provided valuable insights into their accuracy and efficiency. For instance, the simpler logistic regression model performs on par with more complex and costly transformer models when analyzing emoji-only data. We also critically note that many Hugging Face models exclude emojis during tokenization, which impacts their effectiveness in contexts where emojis carry significant semantic weight.


\subsection{Implications} %
This study has several significant implications for financial sentiment analysis. Analyzing emoji usage offers valuable insights regarding investor behavior that can be used for models about financial and economic factors. Notably, emoji-based sentiment analysis, which requires fewer data and computational resources than traditional text-based methods, is particularly promising for high-frequency trading (where even microseconds matter~\cite{aquilina2022quantifying}) and other scenarios where time, resources, or data may be limited. Moreover, the distinctive emoji usage patterns in financial social media—distinct from those observed in general social media—highlight the need for domain-specific sentiment analysis models. For instance, we observed that current emoji recommendation systems in popular mobile keyboards are not optimized for financial posts, suggesting a potential area for further research and development. Our research encourages the creation of specialized models that take into consideration the unique lexicon, including emojis, used in financial social media. By showing that emojis alone can achieve considerable accuracy in sentiment classification, this study aims to contribute to a paradigm shift in the use of emojis for financial sentiment analysis.

\begin{table}[htbp]
\caption{\rev{Correlation between Frequency of Rockets and Asset Price and Volatility Index Changes.}}
\label{tab:rocket_corr}
\centering
\begin{tabular}{lccc}
\toprule
& QQQ & BTC & VIX \\
\midrule
$\mathrm{Corr}(Rockets,\,\cdot)$ & 0.4962 & 0.5211 & -0.4046 \\
\bottomrule
\end{tabular}
\end{table}

\rev{{\bf Interpretability and Practicality.}}
\rev{The correlations reported in Table~\ref{tab:rocket_corr} serve as a pilot demonstration that emoji-based sentiment signals align with market dynamics. The Rocket Index---constructed as the daily proportion of posts containing the \emoji{rocket} emoji---moves positively with Bitcoin and Nasdaq~100 index returns, and negatively with the CBOE Volatility Index (VIX) index. This pattern confirms intuition: rocket emojis typically express enthusiasm and confidence, which rise during market upswings and decline when volatility or fear (VIX) increases. This interpretability strengthens the practical case for emoji-only approaches: they are transparent, language-agnostic, and computationally lightweight, yet remain relevant to real investor behavior. Such models can thus serve as efficient, explainable proxies for market sentiment in real-time applications.}

\subsection{Limitations}
While this study contributes valuable insights into the use of emojis for financial sentiment analysis, it is important to recognize its limitations. The analysis was confined to data sourced from StockTwits, which, while comprehensive and relevant, does not necessarily represent emoji usage across other platforms where financial discussions occur, such as Twitter or professional financial forums. StockTwits users predominantly communicate in English and come from specific demographic backgrounds, primarily within the United States \cite{Similarweb}. This demographic concentration may not accurately reflect global financial social media, where cultural differences significantly influence emoji usage \cite{guntuku2019studying}. The sentiment analysis conducted was binary, classifying posts as either bullish or bearish. This approach may oversimplify the complex emotions and sentiments expressed in financial discussions, such as uncertainty, caution, and speculation, which are also valuable for understanding market dynamics. It is also useful to note that emojis evolve in their usage and meanings over time, influenced by socio-cultural and technological changes \cite{robertson2021semantic}. 

\rev{Emoji-based sentiment analysis also faces an information challenge: emojis can be used sarcastically, ironically, or in combinations whose meaning depends on conversational context. Such nuances are hard to capture with emoji-only models and likely explain why text–emoji models perform better.} 

\rev{Another limitation is potential community bias. StockTwits activity is mostly driven by retail investors and sub-communities such as speculative meme-stock traders. Their posting styles differ from those of institutional or other investors. As a result, the model may overrepresent certain emojis. This effect is mitigated by including a diverse set of assets---including equities from the S\&P\,500, major cryptocurrencies, and the gold index---which together reflect the major discussions on the platform. However, the learned patterns are still valuable for the community they represent, which forms a significant part of online financial discussion. Prior studies show that emoji use varies across communities and cultures~\cite{guntuku2019studying, demirel2024text}, suggesting that such sub-cultural variation is normal in social media rather than unique to our dataset.}

\rev{From an ethical standpoint, systems built only on emoji-based sentiment could be manipulated. For example, coordinated groups might flood discussions with specific emojis to create artificial optimism or pessimism. Future work should explore ways to detect such adversarial behaviors and adjust sentiment models to reduce the impact of such manipulation.}


\subsection{Future Work}
There are several promising areas of future research related to the topic of this study. One potential area of investigation is the creation of emoji-based financial sentiment indices, which could offer efficient and interpretable insights into market sentiment. Additionally, moving beyond binary sentiment classification, future research could explore the use of emojis for multinomial sentiment analysis to capture a broader spectrum of emotions and states relevant to financial decision-making, such as fear, greed, hope, rationality, and uncertainty. This approach would provide a more nuanced understanding of the emotions of financial markets, recognizing that investors with different states may impact the market differently, even if their overall sentiment (bullish or bearish) appears similar.

A formal study to establish that emojis in financial social media are culturally unbiased and, thus, can be used to help NLP models to generalize across different languages is also a very interesting research direction---albeit, the data at hand indicates a strong imbalance toward English on the StockTwits platform.

Beyond sentiment detection, emojis also hold the potential for identifying complex linguistic structures, such as sarcasm in financial communications, or for detecting spam. Moreover, investigating how emoji usage correlates with post-engagement could yield insights into the ways emojis influence investor interactions.

While our study demonstrates the effectiveness of emojis in sentiment analysis, there remains room for improvement in both sentiment analysis models and emoji lexicons. Future research can attempt to improve the accuracy or efficiency of financial sentiment analysis models. By sharing our code and methodology, we aim to facilitate reproducibility and encourage further research in this domain. We believe this can enable researchers to build upon our findings and improve the accuracy of financial sentiment analysis models, particularly given the limitations observed in previous studies. Additionally, there is a significant opportunity for future research to compare the effectiveness of different financial sentiment analysis models and emoji lexicons.


\rev{
\section{Ethical Compliance} \label{sec:ethical_compliance} All data used in this study were collected from publicly accessible StockTwits messages. The dataset contains only publicly posted content and does not include any private or direct messages. We process the data solely for academic research purposes and focus on aggregate sentiment patterns, without attempting to infer emotions or behavior of individual users. Under the European Union's General Data Protection Regulation's (GPDR)~\cite{GDPR} 
academic-research exemption (Article 85, pp.\ 16--17), we process only publicly available information for scholarly purposes, and---since no private user data is involved---our procedures fall under the GDPR's legitimate‑interest basis (Article 6(1)(f), p.\ 12). In line with the GDPR principle of data minimisation, we restrict our analysis and discard any material not essential to our research objectives.
}

\section{Conclusions}
\label{sec: conclusions}

Our research demonstrates that emojis, often ignored in previous work and removed during preprocessing, hold significant value in financial sentiment analysis. Our findings suggest that emoji-based sentiment classification is not only effective but also computationally efficient, requiring significantly less data and processing power compared to text-based models. 

The results show that models incorporating both text and emojis achieved the highest classification performance, with an F$_1$ score of 0.88 \rev{on a balanced dataset of 528{,}000 StockTwits posts (evenly split between bullish and bearish)}. \rev{Emoji-only models achieved F$_1 \approx 0.75$, capturing most of the predictive signal of the combined models while being substantially more computationally efficient.} This efficiency makes them particularly attractive for time-sensitive financial applications such as high-frequency trading, where even microseconds matter \cite{aquilina2022quantifying}. Furthermore, our results indicate that even small training datasets (as few as 1,000 labeled posts) can train accurate sentiment classifiers when using emojis, whereas text-based models require significantly larger datasets to achieve comparable accuracy.

The emoji sentiment lexicon developed in this paper also provides insights into why emojis achieve such efficiency---certain emojis and emoji pairs alone can predict sentiment with over 90\% accuracy. 
Another key insight from our analysis is that simpler models, such as logistic regression, achieve comparable accuracy to more complex transformer-based models when trained on emoji-only data. This indicates that emojis, as compact sentiment indicators, reduce the need for sophisticated deep learning architectures in financial sentiment analysis. 
Finally, our comparative analysis of emoji usage between financial and general social media contexts confirmed that financial platforms exhibit distinct emoji patterns, underscoring the necessity of domain-specific sentiment analysis models.


\rev{\section*{Reproducibility and Data Availability}
Code and data are publicly available at  
\url{https://github.com/AhmedMahrous00/finmoji_replication}.}

\section*{Acknowledgments}

We would like to thank the anonymous reviewers for their valuable comments and suggestions.

The research reported in this publication was supported by funding
from King Abdullah University of Science and Technology (KAUST)
- Center of Excellence for Generative AI - under award number 5940.

\input{emojiPaper.bbl}

\end{document}